\documentclass[10pt,twocolumn,letterpaper]{article}

\usepackage{iccv}
\usepackage{algorithm}
\usepackage{algpseudocode}
\usepackage{times}
\usepackage{epsfig}
\usepackage{graphicx}
\usepackage{amsmath}
\usepackage{amssymb}
\usepackage{bm}
\usepackage{multirow}
\usepackage{makecell}
\usepackage{pifont}
\usepackage{footmisc} 

\usepackage[pagebackref=true,breaklinks=true,letterpaper=true,colorlinks,bookmarks=false]{hyperref}

\iccvfinalcopy 

\def\iccvPaperID{6280} 

\ificcvfinal\fi

\begin{document}
	
	\title{DDS2M:\ Self-Supervised Denoising Diffusion Spatio-Spectral Model for Hyperspectral Image Restoration}
	
	\author{Yuchun Miao$^{1}$ \and Lefei Zhang$^{1}$\thanks{Corresponding Author} \and Liangpei Zhang$^{2}$ \and Dacheng Tao$^{3,4}$ \and \\
		$^1$School of Computer Science, Wuhan University\\
		$^2$ State Key Lab. of Information Engineering in Surveying, Mapping and Remote Sensing, Wuhan University\\
		$^3$JD Explore Academy\\
    	$^4$School of Computer Science, University of Sydney\\
		{\tt\small \{miaoyuchun, zhanglefei, zlp62\}@whu.edu.cn, dacheng.tao@gmail.com}
	}

\maketitle
\ificcvfinal\thispagestyle{empty}\fi

\begin{abstract}
	Diffusion models have recently received a surge of interest due to their impressive performance for image restoration, especially in terms of noise robustness. However, existing diffusion-based methods are trained on a large amount of training data and perform very well in-distribution, but can be quite susceptible to distribution shift. This is especially inappropriate for data-starved hyperspectral image (HSI) restoration. To tackle this problem, this work puts forth a self-supervised diffusion model for HSI restoration, namely Denoising Diffusion Spatio-Spectral Model (\texttt{DDS2M}), which works by inferring the parameters of the proposed Variational Spatio-Spectral Module (VS2M) during the reverse diffusion process, solely using the degraded HSI without any extra training data. In VS2M, a variational inference-based loss function is customized to enable the untrained spatial and spectral networks to learn the posterior distribution, which serves as the transitions of the sampling chain to help reverse the diffusion process. Benefiting from its self-supervised nature and the diffusion process, \texttt{DDS2M} enjoys stronger generalization ability to various HSIs compared to existing diffusion-based methods and superior robustness to noise compared to existing HSI restoration methods. Extensive experiments on HSI denoising, noisy HSI completion and super-resolution on a variety of HSIs demonstrate \texttt{DDS2M}'s superiority over the existing task-specific state-of-the-arts.

\end{abstract}

\section{Introduction}

\begin{figure}[t!] 
	\centering
	\includegraphics[scale=0.35]{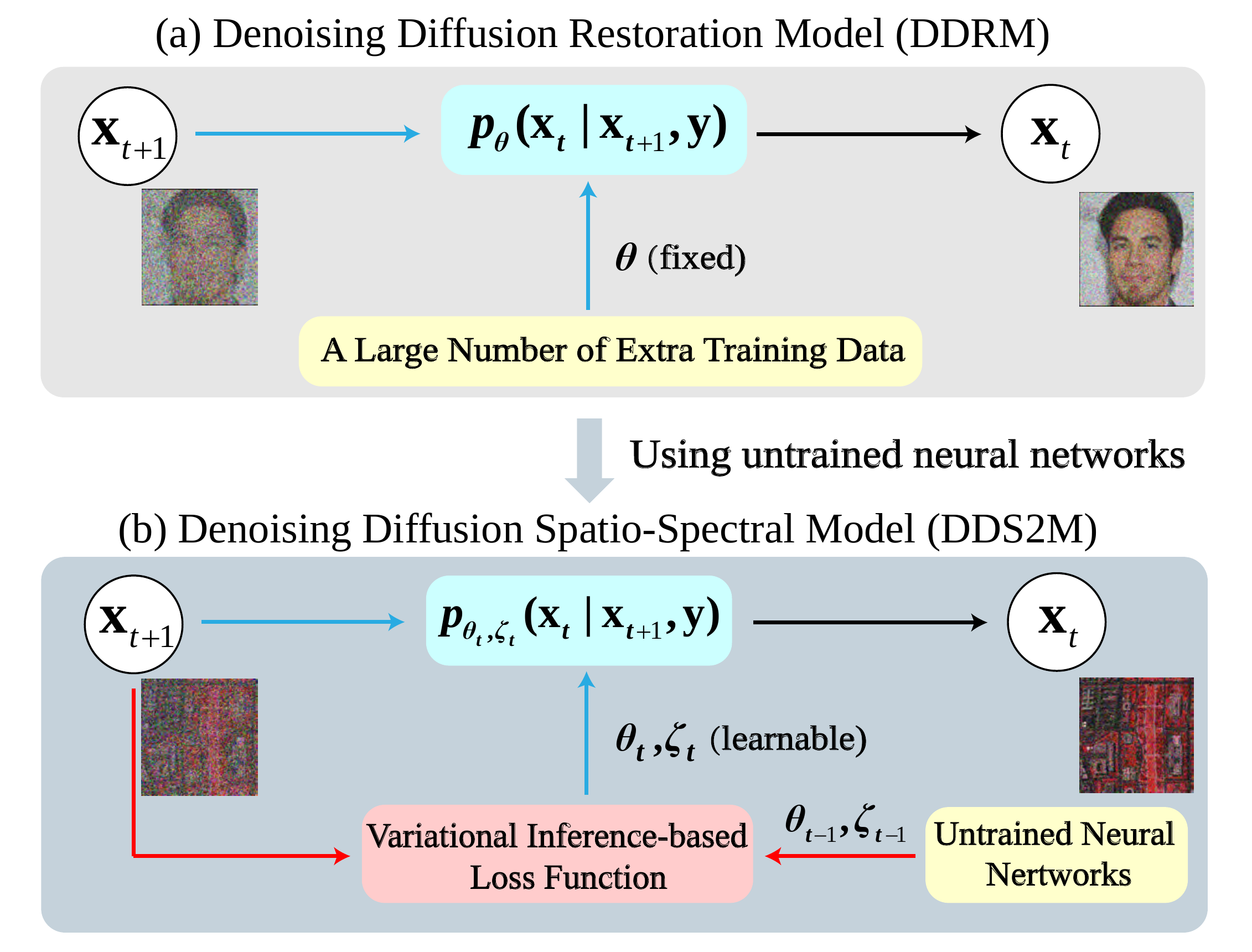}
	\caption{Comparison between \texttt{DDRM} and our self-supervised \texttt{DDS2M}. (a) \texttt{DDRM} utilizes a denoising network pre-trained on a large number of extra training data to reverse the diffusion process. (b) Our \texttt{DDS2M} works by inferring the untrained neural networks' parameters $\{\bm{\theta},\bm{\zeta}\}$ during the reverse diffusion process, only using the degraded HSI $\mathbf{y}$ without any extra training data. The untrained neural networks and the variational inference-based loss function constitute the proposed Variational Spatio-Spectral Module (VS2M).\label{fig:motivation1}}
\end{figure}


As a new trendy generative model, diffusion models~\cite{sohl2015deep,ho2020denoising,nichol2021improved,song2021denoising} have attracted significant attention in the community owing to their state-of-the-art performance in image synthesis~\cite{dhariwal2021diffusion}. In essence, diffusion model is a parameterized sampling chain trained using a variational bound objective, which is equivalent to that of score-based models~\cite{song2019generative,song2020improved,song2021scorebased}. After training, samples are generated by the sampling chain, starting from white noise and gradually denoising to a clean image.

Remarkably, diffusion models can go beyond image synthesis~\cite{gu2022vector,rombach2022high,liu2023more}, and have been widely utilized in image restoration tasks, such as super-resolution~\cite{kawar2022denoising,wang2022zero,saharia2021image,choi2021ilvr}, inpainting~\cite{kawar2022denoising,wang2022zero,saharia2022palette,lugmayr2022repaint,song2019generative,song2021scorebased}, denoising~\cite{kawar2022denoising}, and so on. Among these methods, \texttt{DDRM}~\cite{kawar2022denoising}, a diffusion-based image restoration framework, has achieved powerful robustness to noise, which is also noteworthy for hyperspectral images (HSIs). HSIs often suffer from noise corruption due to the limited light, photon effects, and atmospheric interference~\cite{li2023spatial}. This motivates us to inherit the powerful noise robustness of \texttt{DDRM}~\cite{kawar2022denoising} to HSI restoration by capitalizing on the power of diffusion model for HSI restoration.


However, harnessing the power of the diffusion model for HSI restoration is challenging. The bottleneck lies in the poor generalization ability to HSIs in various scenarios. Existing diffusion-based methods are excessively dependent on the adversity and quantity of the training data, and often focus on a specific domain, such as the face. As a result, these methods may perform very well in-distribution, but can be quite susceptible to distribution shifts, resulting in degraded performance. This is particularly inappropriate for data-poor applications such as HSI restoration, where very limited HSIs are available for training~\cite{miao2021hyperspectral}. This is because HSIs are much more expensive to acquire in real-world scenarios, compared to natural RGB images. In addition, different sensors often admit large different specifications, such as the frequency band used, the spatial and spectral resolution. Therefore, a diffusion model trained on HSIs captured by one sensor may not be useful for HSIs captured by other sensors. In addition to the generalization ability issues mentioned above, how to leverage the intrinsic structure of HSIs is also critical for harnessing the power of the diffusion model for HSI restoration. Bearing the above concerns in mind, an effective diffusion model tailored for HSI restoration, which is able to generalize to HSIs in various practical scenarios and leverage the intrinsic structure of HSIs, is highly desired.  



To address the generalization ability problem mentioned above, one remedy is to use the emerging untrained neural networks, such as those in \cite{ulyanov2018deep,sidorov2019deep,gandelsman2019double}. These methods learn a generative neural network directly from a single degraded image, rather than from a large volume of external training data. The rationale is that an appropriate neural network architecture, without training data, could already encode much critical low-level image statistical prior information. Owing to their training data-independent nature, untrained networks can usually generalize well to the wild data. Meanwhile, due to our need to flexibly cope with various HSIs in real scenarios, untrained networks are rendered as a natural choice. In addition, their powerful expressiveness allows the deployment of such untrained networks in the diffusion models for HSI restoration.

In this work, we put force a self-supervised Denoising Diffusion Spatio-Spectral Model (\texttt{DDS2M}), which can cleverly alleviate the generalization ability problem, while exploiting the intrinsic structure information of the underlying HSIs. \texttt{DDS2M} is a denoising diffusion generative model that progressively and stochastically denoises samples into restored results conditioned on the degraded HSI and the degradation model after a finite time. Unlike existing diffusion models~\cite{sohl2015deep,ho2020denoising,song2021denoising,kawar2022denoising,wang2022zero}, which use a neural network pre-trained a large number of training data, \texttt{DDS2M} reverses the diffusion process by virtue of the proposed Variational Spatio-Spectral Module (VS2M), solely using the degraded HSI without any extra training data; see Figure \ref{fig:motivation1} for visual comparison with \texttt{DDRM}~\cite{kawar2022denoising}. 


Specifically, the proposed VS2M consists of two types of untrained networks (i.e., untrained spatial and spectral networks) and a customized variational inference-based loss function. The untrained spatial and spectral networks leverage the intrinsic structure of HSIs by modeling the abundance maps and endmembers derived from the linear mixture model~\cite{bioucas2012hyperspectral}, respectively. The variational inference-based loss function is customized to enable these untrained networks to learn the posterior distribution of the task at hand. The specific contributions of this work are summarized as follows:

\smallskip  


\noindent
$\bullet$
We propose a self-supervised Deep Diffusion Spatio-Spectral Model (\texttt{DDS2M}). Benefiting from its diffusion process and self-supervised nature, \texttt{DDS2M} enjoys stronger robustness to noise relative to existing HSI restoration methods and superior generalization ability to various HSIs relative to existing diffusion-based methods. To the best of our knowledge, \texttt{DDS2M} is the first self-supervised diffusion model that can restore HSI only using the degraded HSI without any additional training data. 

\noindent
$\bullet$ 
We design a variational spatio-spectral module (VS2M) to help reverse the diffusion process, which serves as the transitions of the sampling chain. VS2M is capable of approximating the posterior distribution of the task at hand by leveraging the intrinsic structure of the underlying HSI.


\noindent
$\bullet$ 
Extensive experiments on HSI denoising, noisy HSI completion and super-resolution illustrate the superiority of \texttt{DDS2M} over the existing task-specific state-of-the-arts, especially in terms of the robustness to noise, and the generalization ability to HSIs in diverse scenarios.



\begin{figure*}[t!] 
	\centering
	\includegraphics[scale=0.45]{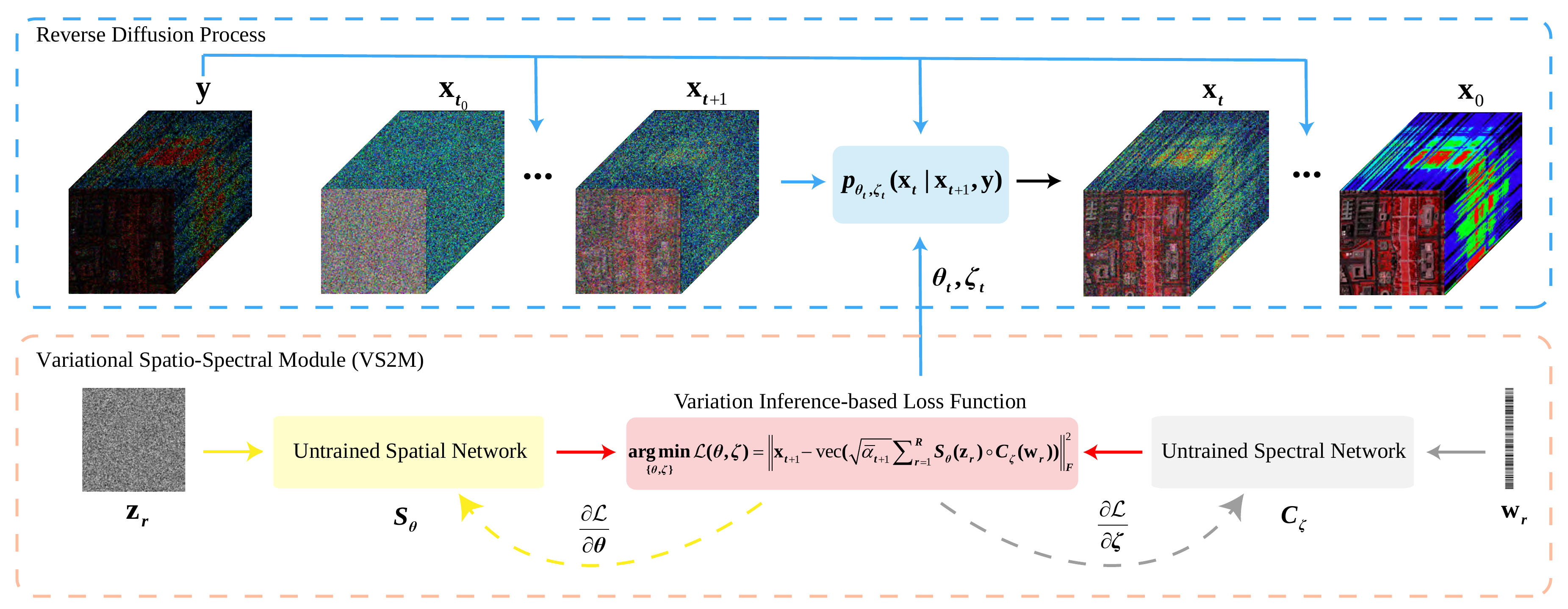}
	\caption{An overview of the proposed self-supervised \texttt{DDS2M}. In \texttt{DDS2M}, the diffusion process is reversed with the help of the proposed VS2M, solely using the degraded HSI without any extra training data. VS2M consists of the untrained spatial and spectral networks (aiming at leveraging the intrinsic structure of HSIs) and the variational inference-based loss function (aiming at enabling the untrained networks to learn the posterior distribution). \label{fig:framework}}
\end{figure*}

\section{Related Works}
\subsection{HSI Restoration Methods}\label{sec:hsi_restoration_method}
HSI restoration is a long-standing problem with a wide range of applications, with model-based approaches dominating the early years~\cite{zhang2013hyperspectral,xiong2019hyperspectral,yuan2019tensor}. Recently, triggered by the expressive power of deep neural networks, a plethora of supervised~\cite{bodrito2021trainable,gong2022learning,wang2021hyperspectral} and self-supervised methods~\cite{sidorov2019deep,miao2021hyperspectral} were developed.



The supervised methods mainly concentrate on exploring different neural network architectures to learn a mapping from a degraded HSI to the ground truth, such as convolution neural network~\cite{maffei2019single, yuan2018hyperspectral}, recurrent neural network~\cite{fu2021bidirectional}, and transformer~\cite{chen2022hider,li2023spatial}. The main bottleneck of these supervised methods is that their performance is limited by the adversity and amount of training data, and is often susceptible to distribution outliers. In contrast, our \texttt{DDS2M} is not affected by such distribution outliers, since no extra training data is required in \texttt{DDS2M}.



Among the self-supervised methods, a representative family is the untrained neural network-based methods~\cite{sidorov2019deep,miao2021hyperspectral}. As a promising tool for image restoration, untrained neural networks enjoy the expressive power of neural networks yet do not require additional training data~\cite{ulyanov2018deep}. Ulyanov \textit{et al.}~\cite{ulyanov2018deep} first extended untrained neural network from RGB images to HSIs, putting forth a self-supervised HSI restoration framework. Then, Luo \textit{et al.}~\cite{luo2021hyperspectral} further proposed a spatio-spectral constrained untrained neural network. Inspired by these methods, Meng \textit{et al.}~\cite{meng2021self} integrated untrained neural network into the plug-and-play regime~\cite{zhang2019deep}. In general, these methods learn a generator network directly from the degraded HSI in an iterative scheme. The critical drawback of these methods is that they easily accumulate errors inevitable in the iterative process, being quite fragile to degraded HSI with significant noise. Although our proposed \texttt{DDS2M} is also a multi-step generation process, it does not suffer from such accumulated errors. This is because diffusion-based methods have systematic mathematical formulation, and the errors in the intermediate step can be regarded as noise, which could be refined during the diffusion process~\cite{wang2022sindiffusion}. Therefore, as compared with the above untrained network-based methods, our \texttt{DDS2M} is able to decently restore high-quality HSIs from the degraded HSI corrupted by noise.


\subsection{Diffusion Models for Image Restoration}\label{sec:diffusion_model}
Recent emerged diffusion models have been widely utilized in image restoration. One branch of these works mainly focuses on tailoring a diffusion model suitable for a specific task, often leading to remarkable performance at the expense of flexibility across different tasks; see~\cite{saharia2021image,lugmayr2022repaint,whang2022deblurring}. Another branch is concerned with tailoring a diffusion model that can be flexibly applied to different tasks; see~\cite{kawar2022denoising,wang2022zero,saharia2022palette}. To achieve this, these methods leave the training procedure intact, and only modify the inference procedure so that one can sample the restored image from a conditional distribution related to the task at hand. Among them, a representative method is \texttt{DDRM}~\cite{kawar2022denoising}, which achieves promising performance in multiple useful scenarios, including denoising, noisy super-resolution, and noisy completion, especially in terms of the robustness to noise. 

However, the main shortcoming of these diffusion-based methods is their generalization ability to the wild data. These methods excessively depend on the adversity and amount of training data, and may perform very well in-distribution, but can be quite susceptible to distribution shifts, sometimes resulting in severely degraded performance. This becomes more problematic for data-poor applications such as HSI restoration. In this work, we aim to inherit the advantage of diffusion model (i.e., noise robustness) to HSI restoration, and boost its generalization ability to HSIs in practical scenarios.

\section{Notations and Preliminaries}
\subsection{Notations}
A scalar, a vector, a matrix, and a tensor are denoted as $x$, $\bf{x}$, $\bf{X}$, and $\mathcal{X}$, respectively. ${\bf x}^{(i)}$, ${\bf X}^{(i,j)}$, and ${\mathcal X}^{(i,j,k)}$ denote the $i$-th, $(i,j)$-th, and $(i,j,k)$-th element of $\mathbf{x}\in \mathbb{R}^{I}$, $\mathbf{X}\in \mathbb{R}^{I\times J}$, and $\mathcal{X}\in \mathbb{R}^{I\times J\times K}$, respectively. The Frobenius norms of $\bf{x}$ are denoted as $\|\bf{x}\|_{F}=\sqrt{\sum_{i}\bf{x}^{(i)}\bf{x}^{(i)}}$. Given $\mathbf{y}\in \mathbb{R}^{N}$ and a matrix $\mathbf{X}\in \mathbb{R}^{I\times J}$, the outer product is defined as $\mathbf{X}\circ \mathbf{y}$. In particular, $\mathbf X\circ \mathbf y \in \mathbb{R}^{I\times J\times N}$ and $(\mathbf{X} \circ \mathbf{y})^{(i,j,n)} = \mathbf{X}^{(i,j)}\mathbf{y}^{(n)}$. The ${\rm vec}(\mathbf X)$ operator represents ${\rm vec}(\mathbf{X})=[\mathbf{X}^{(:,1)};\ldots;\mathbf{X}^{(:,J)}] \in \mathbb{R}^{IJ}$, and ${\rm vec}(\mathcal X)$ is further defined as ${\rm vec}(\mathcal{X}) = [{\rm vec}(\mathcal{X}^{(:,:,1)});\ldots;{\rm vec}(\mathcal{X}^{(:,:,K)})] \in \mathbb{R}^{IJK}$. 

\subsection{Degradation Model}
The goal of HSI restoration is to recover a HSI from potentially noisy degraded HSI given through a known linear degradation model. In general, HSI restoration can be formulated as
\begin{equation}\label{eq:general}
	\mathbf{y} = \mathbf{H}\mathbf{x} + \mathbf{z},
\end{equation}
\noindent
where $ \mathbf{x}  \in \mathbb{R}^{n} $ is the vector version of the original HSI $ \mathcal{X} $ defined as $ \mathbf{x} = {\rm vec(\mathcal{X})}$, $ \mathbf{y}  \in \mathbb{R}^{m} $ is corresponding to the degraded HSI $ \mathcal{Y} $ defined as $ \mathbf{y} = {\rm vec}(\mathcal{Y})$, $\mathbf{H}$  is the degradation matrix that depends on the restoration task at hand, and $\mathbf{z} \sim \mathcal{N}\left(0, \sigma_{\mathbf{y}}^2 \boldsymbol{I}\right)$ represents an \textit{i.i.d.} additive Gaussian noise with standard deviation $\sigma_{\mathbf{y}}$. It is worth noting that in this work, following previous diffusion-based methods~\cite{kawar2022denoising,ho2020denoising,song2021denoising,lugmayr2022repaint,saharia2021image,whang2022deblurring}, $ \mathbf{x} $ and $\mathbf{y}$ in Eqn. \eqref{eq:general} are all scaled linearly to the range of $[-1, 1]$, which ensures the neural network to operate on consistently scaled inputs during the reverse diffusion process. Therefore, when they are linearly scaled back to the range of $[0, 1]$, the standard deviation of the Gaussian noise becomes $\sigma=0.5\sigma_y$.


\section{Denoising Diffusion Spatio-Spectral Models}
In this section, we introduce the proposed \texttt{DDS2M}. The key idea behind \texttt{DDS2M} is to reverse the diffusion process solely using the degraded HSI without extra training data, with the help of the proposed VS2M. We first give an introduction to the diffusion process for image restoration, then describe our design in VS2M, and finally elaborate on the VS2M-aided reverse diffusion process.


\subsection{Diffusion Process for Image Restoration}
Diffusion models for image restoration are generative models with Markov chain $\mathbf{x}_T \rightarrow \mathbf{x}_{T-1} \rightarrow \ldots \rightarrow \mathbf{x}_1 \rightarrow \mathbf{x}_0$ conditioned on $ \mathbf{y} $~\cite{kawar2022denoising}, which has the following marginal distribution equivalent to that in~\cite{ho2020denoising,song2021denoising}: 
\begin{equation}\label{eqn:marginal_distribution}
	q\left(\mathbf{x}_t | \mathbf{x}_0\right)=\mathcal{N}\left(\mathbf{x}_t ; \sqrt{\bar{\alpha}_t} \mathbf{x}_0,\left(1-\bar{\alpha}_t\right) \mathbf{I}\right) 
\end{equation}
with
\begin{equation}
	\alpha_t=1-\beta_t, \quad \bar{\alpha}_t=\prod_{i=0}^t \alpha_i,
\end{equation}
\noindent
where $ \mathbf{x}_0 $ and $ \mathbf{y} $ are the vector version of high-quality HSI $ \mathcal{X} $ and degraded  HSI $ \mathcal{Y} $, and $ \beta_t $ is a hyperparameter. The \textit{forward process} (i.e., \textit{diffusion process} ) progressively injects Gaussian noise to the original data $\mathbf{x}_0$ and obtains $\mathbf{x}_T$ that looks indistinguishable from pure Gaussian noise, while the \textit{reverse diffusion process} samples a slightly less noisy image $\mathbf{x}_{t}$ from $\mathbf{x}_{t+1}$ by leveraging the forward process posterior distribution $q\left(\mathbf{x}_{t} | \mathbf{x}_{t+1}, \mathbf{x}_{0}, \mathbf{y}\right)$. More details can be found in the supplementary materials. 

In \texttt{DDRM}, denoising is performed using a network pre-trained on a large number of additional training data like other diffusion models~\cite{kawar2022denoising,ho2020denoising,song2021denoising,lugmayr2022repaint,saharia2021image,whang2022deblurring}, which perform well in-distribution, and can be susceptible to distribution shift. This is especially inappropriate with data-starved HSI restoration. In this work we break this routine and propose to reverse the diffusion process utilizing the VS2M that can perform denoising solely using the degraded image without any extra training data.

\subsection{Variational Spatio-Spectral Module (VS2M)}\label{sec:optimizing}
The VS2M utilized in \texttt{DDS2M} consists of untrained spatial and spectral networks, and a variational inference-based loss function. The untrained spatial and spectral networks are capable of leveraging the intrinsic structure of HSIs using designated network structures. The variational inference-based loss function is customized to enable these untrained networks to learn the posterior distribution. In this way, the untrained networks and the diffusion model can be incorporated to achieve promising performance.


Under VS2M, HSI $ \mathcal{X} \in\mathbb{R}^{I\times J\times K}$ is represented as:
\begin{equation}\label{eq:LMM}
	\mathcal{X} = \sum\nolimits_{r=1}^R \mathbf{S}_r \circ \mathbf{c}_r,
\end{equation}
where $\mathbf{c}_r\in\mathbb{R}^K$ and $\mathbf{S}_r\in\mathbb{R}^{I\times J}$  represent the $r$-th endmember and the $r$-th endmember's abundance map, respectively, and $R$ is the number of endmembers contained in the HSI. More details about the decomposition in Eqn. \eqref{eq:LMM} can be found in the supplementary materials. Here we introduce the untrained network architecture and the variational inference-based loss function individually.

\noindent
\textbf{Untrained Network Architecture.}
The physical interpretation of  $\mathbf{S}_r$ and $\mathbf{c}_r$ makes it possible to utilize certain untrained networks to model these factors. Specifically, untrained U-Net-like ``hourglass`` architecture in~\cite{ulyanov2018deep} and untrained full-connected networks (FCNs) are employed for abundance map modeling and endmember modeling, since abundance maps reveal similar qualities of the nature images~\cite{qian2016matrix} and the endmembers can be regarded as relatively simple 1D signals, as was done in~\cite{miao2021hyperspectral}. Following this perspective, we model the HSI $\mathbf{x}\in\mathbb{R}^{IJK}$  as follows:\footnote{Actually, The parameters of the $ r $ U-Nets are independent of each other, as are the parameters of the $ r $ FCNs. In order to simplify notations, here we use $\bm \theta$ and $\bm \zeta$ to represent $\{\theta_r\}_{r=1}^{R} $ and $\{\zeta_r\}_{r=1}^{R} $, respectively. } 
\begin{equation}
	{\mathbf{x}} = {\rm vec}(\mathcal{X}) =  {\rm vec}(\sum\nolimits_{r=1}^R {\cal S}_{\mathbf{\bm \theta}}(\mathbf{z}_r) \circ {\cal C}_{\mathbf{\bm \zeta}}(\mathbf{w}_r )),
\end{equation}
where ${\cal S}_{\mathbf{\bm \theta}}(\cdot):\mathbb{R}^{N_a} \rightarrow \mathbb{R}^{I\times J}$ is the untrained U-Net-like network for abundance map generation, and $\bm \theta$ collects all the corresponding network weights; similarly, ${\cal C}_{{\bm \zeta}}(\cdot):\mathbb{R}^{N_s}\rightarrow \mathbb{R}^{K}$ and ${\bm \zeta}$ denote the untrained FCN for endmember generation and the corresponding network weights, respectively; the vectors $\mathbf{z}_r\in\mathbb{R}^{N_a}$ and $\mathbf{w}_r\in\mathbb{R}^{N_s}$ are low-dimensional random vectors that are responsible for generating the $r$-th abundance map and endmember respectively. $ \mathbf{z}_r $ and $ \mathbf{w}_r $ are randomly initialized but fixed during the optimization process. It is worth noting that, instead of directly using the vanilla U-Net structure for abundance map modeling, we propose to introduce the attention mechanism~\cite{woo2018cbam} into the U-Net, which aims to enhance the self-supervised expression ability of the VS2M. The concrete structure of the untrained spatial and spectral networks is illustrated in the supplementary materials.

\noindent
\textbf{Variational Inference-based Loss Function.}
We aim to estimate high-quality HSI $\mathbf{x}_0$  using the aforementioned untrained spatial and spectral networks, and update their parameters at every reverse process step. Denoting $ \{{\bm \theta}_t, {\bm \zeta}_t \}$ as the parameters at step $t$, we first define a learnable generative process $p_{\bm{\theta}_t,\bm{\zeta}_t} \left(\mathbf{x}_{t} | \mathbf{x}_{t+1},\mathbf{y}\right)$ by replacing the $\mathbf{x}_0$ in $q\left(\mathbf{x}_{t} | \mathbf{x}_{t+1}, \mathbf{x}_{0}, \mathbf{x}\right)$ with $ {\mathbf{x}}_{\bm{\theta}_t,\bm{\zeta}_t} $, i.e.,  
\begin{equation}\label{eqn:postior}
	p_{\bm{\theta}_t,\bm{\zeta}_t} \left(\mathbf{x}_{t} | \mathbf{x}_{t+1},\mathbf{y}\right) \triangleq q\left(\mathbf{x}_{t} | \mathbf{x}_{t+1}, {\mathbf{x}}_{\bm{\theta}_t,\bm{\zeta}_t}, \mathbf{y}\right),
\end{equation}
\noindent
where  $ {\mathbf{x}}_{\bm{\theta}_t,\bm{\zeta}_t} $ denotes the vector version of the estimated HSI at reverse process step $t$, i.e.,
\begin{equation}\label{eqn:ds2p}
	{\mathbf{x}}_{\bm{\theta}_t,\bm{\zeta}_t} = {\rm vec}(\sum\nolimits_{r=1}^R {\cal S}_{\bm{\theta}_t}(\mathbf{z}_r) \circ {\cal C}_{\bm{\zeta}_t}(\mathbf{w}_r ))
\end{equation} 
The goal of \texttt{DDS2M} is to find a set of parameters $ \{{\bm \theta}_t, {\bm \zeta}_t\} $ to make $p_{\bm{\theta}_t,\bm{\zeta}_t} \left(\mathbf{x}_{t} | \mathbf{x}_{t+1},\mathbf{y}\right)$ as close to $ q\left(\mathbf{x}_{t} | \mathbf{x}_{t+1}, \mathbf{x}_0,\mathbf{y}\right) $ as possible, by maximizing the 
variational lower bound of the log likelihood objective:
\begin{equation}\label{eqn:objective}
	\begin{aligned}
		 &\mathbb{E}_{q\left(\mathbf{x}_0\right), q\left(\mathbf{y} \mid \mathbf{x}_0\right)}\left[\log p_{{\bm \theta}, {\bm \zeta}}\left(\mathbf{x}_0 | \mathbf{y}\right)\right] \\
		 \ge &\mathbb{E}_{q\left(\mathbf{x}_{0:T}\right), q\left(\mathbf{y}|\mathbf{x}_0\right)}\left[\log p_{{\bm \theta}, {\bm \zeta}}\left(\mathbf{x}_{0:T} | \mathbf{y}\right) - \log q \left(\mathbf{x}_{1:T}|\mathbf{x}_0,\mathbf{y} \right)\right]. 	
	\end{aligned}
\end{equation}
%
Notably, the objective in Eqn.~(\ref{eqn:objective}) can be reduced into a denoising objective, i.e., estimating the underlying high-quality HSI $ \mathbf{x}_0$ from the noisy $ \mathbf{x}_t$ (please refer to the supplementary materials for derivation). By reparameterizing Eqn. \eqref{eqn:marginal_distribution} as 
\begin{equation}
	\mathbf{x}_t\left(\mathbf{x}_0, \boldsymbol{\epsilon}\right)=\sqrt{\bar{\alpha}_t} \mathbf{x}_0+\sqrt{1-\bar{\alpha}_t} \epsilon \ \ \ \ \text { for } \epsilon \sim \mathcal{N}(\mathbf{0}, \mathbf{I}),
\end{equation}
\noindent
our variation inference-based loss function can be designed as follows:
\begin{equation}\label{eqn:proposed_loss}
	\mathop{\arg\min}_{\{{\bm \theta}, {\bm \zeta}\}} ~ \left\| \mathbf{x}_t  - {\rm vec}(\sqrt{\bar{\alpha}_t} \sum\nolimits_{r = 1}^R {{\cal S_{{\bm \theta}}}} ({\mathbf{z}_r}) \circ {\cal C_{{\bm \zeta}}}({\mathbf{w}_r}))\right\|_F^2.
\end{equation}
Intuitively, given a noisy observation $\mathbf{x}_{t+1}$, after optimizing $ \{\bm{\theta}_t,\bm{\zeta}_t \}$ from $\mathbf{x}_{t+1}$ via Eqn.~\eqref{eqn:proposed_loss} using the Adam~\cite{kingma2015adam}, $\mathbf{x}_{\bm{\theta}_t,\bm{\zeta}_t}$ can be derived via Eqn. \eqref{eqn:ds2p}, and then $\mathbf{x}_{t}$ could be sampled from $ p_{\bm{\theta}_t,\bm{\zeta}_t} \left(\mathbf{x}_{t} | \mathbf{x}_{t+1},\mathbf{y}\right)$ defined in Eqn. \eqref{eqn:postior}. In this way, the diffusion process could be reversed in a self-supervised manner with no need for extra training data.


\subsection{VS2M-Aided Reverse Diffusion Process}\label{sec:4.3}
Given a degradation matric $ \mathbf{H}\in\mathbb{R}^{m \times n} $, its singular value decomposition is posed as:
\begin{equation}
	\mathbf{H} = \mathbf{U}\mathbf{\Sigma}\mathbf{V}^\mathsf{T},
\end{equation}
\noindent
where $ \mathbf{U} \in \mathbb{R}^{m \times m} $, $ \mathbf{V} \in \mathbb{R}^{n \times n} $ are orthogonal matrices, and $ \mathbf{\Sigma} \in \mathbb{R}^{m \times n} $ is the rectangular diagonal matrix consisting of the singular values denoted as $s_1 \geq s_2 \geq \ldots \geq s_n$. 
The idea behind this is to tie the noise in the degraded signal $\mathbf{y}$ with the diffusion noise in $\mathbf{x}_{1:T}$, ensuring that the diffusion result $\mathbf{x}_0$ is faithful to the degraded signal $\mathbf{y}$~\cite{kawar2021snips}.

Before illustrating the reverse diffusion process in detail, we first rethink the difference between our \texttt{DDS2M} and other diffusion-based methods~\cite{ho2020denoising,kawar2022denoising} to guide the design of the reverse diffusion process. The main difference is how $\mathbf{x}_0$ is predicted from $\mathbf{x}_t$ at each reverse step. In~\cite{ho2020denoising,kawar2022denoising}, a denoising network is trained on a large amount of additional training data to predict $\mathbf{x}_0$. By exploiting the external prior knowledge, this network could produce satisfactory $\mathbf{x}_0$ even if $\mathbf{x}_t$ looks like pure Gaussian noise. Because of this, such a denoising network could work during the whole reverse diffusion process. However, it is difficult for untrained networks to produce a satisfactory image by denoising an image that is almost pure Gaussian noise. Therefore, starting inference from pure Gaussian is unsuitable for our \texttt{DDS2M}.

Following the above argument, we propose to start inference from a single forward diffusion with better initialization, instead of starting from pure Gaussian noise~\cite{ho2020denoising,kawar2022denoising,nichol2021improved,song2021denoising}. Specifically, we first perturb the degraded HSI $\mathbf{y}$ via the forward diffusion process up to $t_0 < T$, where $t_0$ denotes the step that the reverse diffusion process starts from. 
Denoting $ \bar{\mathbf{x}}^{(i)} $ as the $i$-th index of vector $ \bar{\mathbf{x}}_t=\mathbf{V}^\mathsf{T}\mathbf{x}_t $, $ \bar{\mathbf{y}}^{(i)} $ as the $i$-th index of $ \bar{\mathbf{y}}=\mathbf{\Sigma}^{\dagger}\mathbf{U}^\mathsf{T}\mathbf{y} $,  and $ \bar{\mathbf{x}}_{{\bm \theta}_t, {\bm \zeta}_t}^{(i)} $ as the $ i $-th index of $ \bar{\mathbf{x}}_{{\bm \theta}_t, {\bm \zeta}_t} = \mathbf{V}^\mathsf{T} \mathbf{x}_{{\bm \theta}_t, {\bm \zeta}_t} $, for all $t < t_0$, the variational distribution is defined as:
\begin{equation}\label{eqn:reverse_process1}
	p_{\bm{\theta}_{t_0},\bm{\zeta}_{t_0}} \left(\bar{\mathbf{x}}_{t_0}^{(i)} | \mathbf{y}\right) = \begin{cases}\mathcal{N}(\bar{\mathbf{y}}^{(i)}, \sigma_{t_0}^2-\frac{\sigma_y^2}{s_i^2}) & \text { if } s_i>0 \\ \mathcal{N}(0, \sigma_{t_0}^2) & \text { if } s_i=0\end{cases}
\end{equation}
\begin{equation}\label{eqn:reverse_process2}
	\begin{aligned}
		& p_{\bm{\theta}_{t},\bm{\zeta}_{t}}(\bar{\mathbf{x}}_t^{(i)} | \mathbf{x}_{t+1}, \mathbf{y})= \\
		& \begin{cases}\mathcal{N}(\bar{\mathbf{x}}_{\bm{\theta}_t,\bm{\zeta}_t}^{(i)}+\sqrt{1-\eta^2} \sigma_t \frac{\bar{\mathbf{x}}_{t+1}^{(i)}-\bar{\mathbf{x}}_{\bm{\theta}_t,\bm{\zeta}_t}^{(i)}}{\sigma_{t+1}}, \eta^2 \sigma_t^2) & \text { if } s_i=0 \\
			\mathcal{N}(\bar{\mathbf{x}}_{\bm{\theta}_t,\bm{\zeta}_t}^{(i)}+\sqrt{1-\eta^2} \sigma_t \frac{\bar{\mathbf{y}}^{(i)}-\bar{\mathbf{x}}_{\bm{\theta}_t,\bm{\zeta}_t}^{(i)}}{\sigma_{\mathbf{y}} / s_i}, \eta^2 \sigma_t^2) & \text { if } \sigma_t<\frac{\sigma_{\mathbf{y}}}{s_i} \\
			\mathcal{N}(\left(1-\eta_b\right) \bar{\mathbf{x}}_{\bm{\theta}_t,\bm{\zeta}_t}^{(i)}+\eta_b \bar{\mathbf{y}}^{(i)}, \sigma_t^2-\frac{\sigma_{\mathbf{y}}^2}{s_i^2} \eta_b^2) & \text { if } \sigma_t \geq \frac{\sigma_{\mathbf{y}}}{s_i}\end{cases}
	\end{aligned}
\end{equation}
\noindent
where $\sigma_t$ depending on the hyperparameter $\beta_{1:T}$  denotes the variance of diffusion noise in $\mathbf{x}_t$, and $ \eta$, $\eta_b$ are the hyperparameters, which control the level of noise injected at each timestep. Once $ \bar{\mathbf{x}}_{{\bm \theta}_t, {\bm \zeta}_t} $ is sampled from Eqn. \eqref{eqn:reverse_process2}, it is easy to obtain $ \mathbf{x}_{{\bm \theta}_t, {\bm \zeta}_t} $ exactly by left multiplying $ \mathbf{V} $. And the values of the parameters $ \{\bm \theta_{t_0}, \bm \zeta_{t_0}\} $ are randomly initialized.

It is worth noting that the parameter updating (i.e, $ \{\bm \theta, \bm \zeta\} $) and the reverse diffusion process are iteratively performed. The parameter values of each reverse diffusion step are inherited from the previous step, thus the parameters can be updated continuously during the reverse diffusion process. This reverse diffusion process is summarized in Algorithm~\ref{algo:proposed}.
\begin{algorithm}[h]
	\renewcommand\arraystretch{1.2}
	\caption{Reverse Diffusion Process of \texttt{DDS2M}.}
	\begin{algorithmic}[1]
		\renewcommand{\algorithmicrequire}{\textbf{Input:}} %
		\renewcommand{\algorithmicensure}{\textbf{Output:}}
		\Require The degraded HSI $ \mathbf{y} $, the hyperparameter $R$, $t_0$, $T$, $\beta_{1:t_0}$, $\sigma_{1:t_0}$, $\sigma_y$, $\eta$ and $ \eta_b $.
		\State Randomly initialize $ \bm \theta_{t_0} $, $ \bm \zeta_{t_0} $,$ \mathbf{z}_r $, and $ \mathbf{w}_r $ ;
		\State Obtain $\mathbf{x}_{t_0}$ via reparameterizing Eqn.~\eqref{eqn:reverse_process1};
		\For {$t=t_0-1$ to 1}
		\State Update $\{\bm \theta_{t},\bm \zeta_{t}\}$ via Eqn.~\eqref{eqn:proposed_loss};
		\State Obtain $ {\mathbf{x}}_{\bm{\theta}_t,\bm{\zeta}_t} $ via Eqn. \eqref{eqn:ds2p};
		\State Obtain $\mathbf{x}_{t-1}$ via reparameterizing Eqn.~\eqref{eqn:reverse_process2};
		
		\EndFor
		\Ensure
		The restored HSI $\mathbf{x}_{0}$.
	\end{algorithmic}
	\label{algo:proposed}
\end{algorithm}

\section{Experiments}
\subsection{Comparisons with State-of-the-Arts}
In this paper, our interest lies in inheriting the \texttt{DDRM}'s powerful robustness to noise (which is unavoidable in the hyperspectral imaging process) to HSI restoration. Herein we mainly consider noisy HSI completion, HSI denoising, and noisy HSI super-resolution, and compare the proposed \texttt{DDS2M} with the existing task-specific state-of-the-arts. Two frequently used evaluation metrics, namely, peak signal-to-noise ratio (PSNR) and structure similarity (SSIM), are adopted to evaluate the results. In general, better performance is reflected by higher PSNR and SSIM values. In \texttt{DDS2M}, $T$ is set as \{3000, 1000, 1000\}  for noisy HSI completion, HSI denoising, and noisy HSI super-resolution, respectively, and the step $t_0$ to start reverse the diffusion process is set as $T/2$. We use $\eta=0.95, \eta_b=1$, and linearly increase $\beta_{1:T}$ in which $\beta_1=10^{-4}$ and $\beta_T$ = $\{2\times10^{-3}, 5\times10^{-3}\}$. The variance $\sigma_t$ is set as a constant $\sigma_t = \frac{1-\bar{\alpha}_{t-1}}{1-\bar{\alpha}_t} \beta_t$  for all experiments. The number of endmembers $R$ is selected from $\{5, 10\}$. As for diffusion-based restoration methods \texttt{DDNM}~\cite{wang2022zero} and \texttt{DDRM}~\cite{kawar2022denoising}, the diffusion model in them is trained on a large-scale remote sensing imagery dataset AID~\cite{xia2017aid} containing ten thousands of scene images, and HSI restoration is performed in a channel-by-channel manner. All of the compared methods' parameters are set as suggested by the authors, with parameter fine-tuning efforts to uplift their performance. For implementation details, parameters sensitivity analysis, and inference time analysis, please refer to the supplementary materials. 

\subsubsection{Datasets and Compared Methods}
\noindent
\textbf{Noisy HSI Completion.} The noisy HSI completion aims at recovering the underlying HSI from the noisy incompleted observation. We adopt a wide range of HSIs to conduct the experiments, including 32 natural HSIs\footnote{\url{https://www.cs.columbia.edu/CAVE/databases/multispectral/}} (i.e., \textit{CAVE} dataset~\cite{yasuma2010generalized}), and 3 remote sensing HSIs\footnote{\url{http://lesun.weebly.com/hyperspectral-data-set.html}} (i.e., \textit{WDC Mall}, \textit{Pavia Centre}, and \textit{Pavia University} datasets). The sampling rates are set as \{0.1, 0.2, 0.3\}, and the standard deviation $ \sigma $ of Gaussian noise in the range of [0,1] is set as 0.1. The compared methods consist of seven model-based methods (i.e., \texttt{TMac-TT}~\cite{bengua2017efficient}, \texttt{TNN}~\cite{zhang2014novel}, \texttt{TRLRF}~\cite{yuan2019tensor}, \texttt{FTNN}~\cite{jiang2020framelet}, \texttt{TCTF}~\cite{zhou2017tensor}, \texttt{SN2TNN}~\cite{luo2022self}, and \texttt{HLRTF}~\cite{luo2022hlrtf}), two unsupervised deep learning-based methods (i.e., \texttt{DIP2D}~\cite{sidorov2019deep} and \texttt{DIP3D}~\cite{sidorov2019deep}), and two diffusion-based methods (i.e., \texttt{DDRM}~\cite{kawar2022denoising} and \texttt{DDNM}~\cite{wang2022zero}).

\noindent
\textbf{HSI Denoising.} The HSI denoising aims at recovering the clean HSI from its noisy observation. The data adopted in this experiment is the same as that in HSI completion, including 32 natural HSIs and 3 remote sensing HSIs. Herein we mainly consider Gaussian noise, and the standard deviation of Gaussian noise $ \sigma $ in the range of $[0,1]$ is set as $\{0.1, 0.2, 0.3\}$. The compared methods consist of six model-based methods (i.e., \texttt{LRMR}~\cite{zhang2013hyperspectral}, \texttt{LRTDTV}~\cite{wang2017hyperspectral}, \texttt{LRTFL0}~\cite{xiong2019hyperspectral}, \texttt{E3DTV}~\cite{peng2020enhanced}, \texttt{HLRTF}~\cite{luo2022hlrtf}, and \texttt{NGMeet}~\cite{he2019non}), two unsupervised deep learning-based methods (i.e., \texttt{DIP2D}~\cite{sidorov2019deep} and \texttt{DIP3D}~\cite{sidorov2019deep}), and a supervised deep learning-based method (i.e., \texttt{SST}~\cite{li2023spatial}). Since the purpose of the comparison with supervised methods in this work is to highlight the generalization ability of our methods, we directly use the models of \texttt{SST} trained on ICVL~\cite{arad2016sparse} with Gaussian noise provided by the authors.

\noindent
\textbf{Noisy HSI Super-Resolution.} The noisy HSI super-resolution aims at recovering high-resolution HSI from its noisy low-resolution counterpart. We adopt CAVE dataset~\cite{yasuma2010generalized} to conduct the experiments. The scale factor is set as $\times$2, $\times$4, and $\times$8, and the standard deviation of Gaussian noise $\sigma$ in the range of $[0,1]$ is set as 0.1. The compared methods include three supervised deep learning-based methods (i.e., \texttt{SFCSR}~\cite{wang2020hyper}, \texttt{RFSR}~\cite{wang2021hyperspectral}, and \texttt{SSPSR}~\cite{jiang2020learning}), a model-based method (i.e., \texttt{LRTV}~\cite{shi2015lrtv}), two unsupervised deep learning-based methods (i.e., \texttt{DIP2D}~\cite{sidorov2019deep} and \texttt{DIP3D}~\cite{sidorov2019deep}), and a diffusion-based method (i.e., \texttt{DDRM}~\cite{kawar2022denoising}). In order to comprehensively compare with supervised methods in terms of generalization ability to other datasets and other noise standard deviations, we train each supervised model under five different settings, i.e., CAVE without noise denoted as \texttt{xxx(0)}, CAVE with 0.1 Gaussian noise denoted as \texttt{xxx(0.1)}, CAVE  with 0.05 Gaussian noise denoted as \texttt{xxx(0.05)}, CAVE with 0.03 Gaussian noise denoted as \texttt{xxx(0.03)}, and Chikusei dataset~\cite{yokoya2016airborne} with 0.1 Gaussian noise denoted as \texttt{xxx(0.1)}*. Here \texttt{xxx} denotes the method name, i.e., \texttt{SFCSR}, \texttt{RFSR}, and \texttt{SSPSR}.


\subsubsection{Experimental Results}

\vspace{-0.1cm}
\begin{table}[!htbp]
	\scriptsize
	\setlength{\tabcolsep}{2.2pt}
	\renewcommand\arraystretch{1.3}
	\caption{The average quantitative results for noisy HSI completion. The \textbf{best} and \underline{second-best} values are highlighted.}
	\centering
	\begin{tabular}{clcccccc}
		\Xhline{1.0pt}
		\multicolumn{2}{c}{Sampling Rate} & \multicolumn{2}{c}{0.1} &  \multicolumn{2}{c}{0.2} & \multicolumn{2}{c}{0.3} \\ \hline
		Dataset & Method &\; PSNR & SSIM &\; PSNR & SSIM &\; PSNR & SSIM \\ \hline
		~ & TNN &\; 23.841  & 0.334  &\; 24.241  & 0.333  &\; 24.361  & 0.333  \\ 
		~ & TMac-TT &\; 21.516  & 0.473  &\; 21.104  & 0.439  &\; 21.501  & 0.407  \\ 
		~& TRLRF &\; 26.745  & 0.548  &\; 27.968  & 0.626  &\; 28.427  & 0.655  \\ 
		~ & DIP2D &\; 28.621 & 0.676  &\; 29.412  & 0.693  &\; 29.971  & 0.704  \\ 
		\textbf{Natural HSI}  & DIP3D &\; 24.938  & 0.592  &\; 25.273  & 0.603  &\; 25.342  & 0.606  \\
		CAVE Dataset & FTNN &\; 25.071  & 0.459  &\; 26.293  & 0.495  &\; 26.923  & 0.515  \\ 
		consists of 32 HSIs & FCTN &\; 26.778  & 0.578  &\; 27.547  & 0.631  &\; 27.812  & 0.649  \\
		each with a size of & SN2TNN &\; 25.883  & 0.532  &\; 27.236  & 0.585  &\; 28.101  & 0.617  \\ 
		$256\times256\times31$ & HLRTF &\; \underline{29.514}  & \underline{0.700}  &\; \underline{30.076}  & 0.725  &\; 30.728  & 0.748  \\
		~ & DDNM &\; 16.718 & 0.260 &\; 27.847 & 0.607 &\; \underline{32.222} &  0.818 \\ 
		~ & DDRM &\; 24.474  & 0.655  &\; 28.151  & \underline{0.785}  &\; 29.868  & \underline{0.827}  \\
		~ & DDS2M &\; \textbf{32.507}  & \textbf{0.871}  &\; \textbf{34.156}  & \textbf{0.896}  &\; \textbf{35.098}  & \textbf{0.909} \\ \hline
		
		~ & TNN &\; 23.031  & 0.478  &\; 23.030  & 0.488  &\; 22.721  & 0.479  \\ 
		~ & TMac-TT &\; 21.859  & 0.411  &\; 22.026  & 0.417  &\; 21.640  & 0.390  \\ 
		~ & TRLRF &\; 25.402  & 0.644  &\; 25.772  & 0.666  &\; 25.901  & 0.675  \\ 
		\textbf{Remote Sensing HSI} & DIP2D &\; 28.392  & 0.786  &\; 30.600  & 0.857  &\; 31.608  & 0.882  \\ 
		\textit{WDC Mall} & DIP3D &\; 22.204  & 0.399  &\; 22.169  & 0.402  &\; 22.512  & 0.405  \\ 
		$256\times256\times191$ & FTNN &\; 23.956  & 0.523  &\; 25.575  & 0.619  &\; 26.457  & 0.666  \\ 
		\textit{Pavia Centre} & FCTN &\; 24.352  & 0.586  &\; 24.523  & 0.599  &\; 24.591  & 0.604  \\ 
		$256\times256\times87$ & SN2TNN &\; 28.567  & 0.797  &\; 30.513  & 0.848  &\; 31.507  & 0.873  \\ 
		\textit{Pavia University} & HLRTF &\; \underline{29.272}  & \underline{0.825}  &\; \underline{31.001}  & \underline{0.869}  &\; \underline{31.938}  & \underline{0.891}  \\ 
		$192\times192\times80$ & DDNM &\; 21.002 & 0.343 &\; 23.445 & 0.534 &\; 25.758 & 0.657 \\ 
		~ & DDRM &\; 21.423  & 0.371  &\; 23.467  & 0.495  &\; 24.771  & 0.587  \\ 
		~ & DDS2M &\; \textbf{30.277}  & \textbf{0.857}  &\; \textbf{32.179}  &\textbf{ 0.900}  &\; \textbf{33.208}  & \textbf{0.918} \\ \hline
		

	\end{tabular}
	\label{tab:completion}
\end{table}

\begin{table}[!htbp]
	\scriptsize
	\setlength{\tabcolsep}{2.2pt}
	\renewcommand\arraystretch{1.3}
	\caption{The average quantitative results for HSI denoising. The \textbf{best} and \underline{second-best} values are highlighted.}
	\centering
	\begin{tabular}{clcccccc}
		\Xhline{1.0pt}
		\multicolumn{2}{c}{standard deviation} & \multicolumn{2}{c}{0.1} &  \multicolumn{2}{c}{0.2} & \multicolumn{2}{c}{0.3} \\ \hline
		Dataset & Method &\; PSNR & SSIM &\; PSNR & SSIM &\; PSNR & SSIM \\ \hline
		~ & LRMR &\; 30.948  & 0.754  &\; 27.718  & 0.600  &\; 25.698  & 0.496  \\ 
		~ & LRTDTV &\; \underline{37.354}  & \underline{0.937}  &\; 33.598  & 0.863  &\; 30.098  & 0.725  \\ 
		~ & LRTFL0 &\; 34.205  & 0.872  &\; 29.551  & 0.722  &\; 26.155  & 0.641  \\
		
		~ & DIP2D &\; 30.498  & 0.742  &\; 24.663  & 0.605  &\; 20.808  & 0.513  \\ 
		\textbf{Natural HSI} & DIP3D &\; 27.965  & 0.677  &\; 23.759  & 0.559  &\; 20.407  & 0.485  \\
		CAVE Dataset & NGMeet &\; 31.698  & 0.772  &\; 24.964  & 0.621  &\; 20.657  & 0.517  \\ 
		consists of 32 HSIs & E3DTV &\; 33.652  & 0.922  &\; 30.752  & 0.876  &\; 29.044  & \underline{0.836}  \\ 
		each with a size of & HLRTF &\; 37.095  & 0.935  &\; \underline{33.623}  & \underline{0.881}  &\; \underline{31.661}  & \underline{0.836}  \\ 
		$256\times256\times31$  & SST &\; 29.803  & 0.757  &\; 24.519  & 0.627  &\; 20.866  & 0.542  \\ 
		~ & DDNM &\; 29.223 & 0.615 &\; 24.148 & 0.353 &\; 21.104 & 0.226  \\ 
		~ & DDRM &\; 33.391  & 0.895  &\; 29.987  & 0.831  &\; 27.935  & 0.782  \\ 
		~ & DDS2M &\; \textbf{38.021}  & \textbf{0.944}  &\; \textbf{34.879}  & \textbf{0.902}  &\; \textbf{32.951}  & \textbf{0.871} \\ \hline
		
		~ & LRMR &\; 28.223  & 0.838  &\; 26.950  & 0.776  &\; 25.677  & 0.708  \\
		~ & LRTDTV &\; 32.793  & 0.906  &\; 30.017  & 0.835  &\; 28.252  & 0.771  \\ 
		~& LRTFL0 &\; 35.392  & 0.953  &\; \underline{31.907}  & \underline{0.888}  &\; \underline{29.485}  & \underline{0.821}  \\ 
		\textbf{Remote Sensing HSI} & DIP2D &\; 30.991  & 0.872  &\; 27.195  & 0.801  &\; 23.067  & 0.731  \\
		\textit{WDC Mall} & DIP3D &\; 25.973  & 0.625  &\; 24.087  & 0.559  &\; 21.730  & 0.505  \\ 
		$256\times256\times191$ & NGMeet &\; \underline{36.149}  & \underline{0.956}  &\; 28.308  & 0.857  &\; 23.313  & 0.718  \\ 
		\textit{Pavia Centre}  & E3DTV &\; 33.837  & 0.929  &\; 30.167  & 0.850  &\; 28.098  & 0.785  \\ 
		$256\times256\times87$ & HLRTF &\; 34.987  & ~0.932  &\; 31.359  & 0.870  &\; 29.431  & 0.780  \\ 
		\textit{Pavia University} & SST &\; 34.625  & 0.932  &\; 27.487  & 0.820  &\; 22.821  & 0.709  \\ 
		$192\times192\times80$ & DDNM &\; 26.855 & 0.687 &\; 22.433 & 0.439 &\; 19.661 & 0.287 \\
		~ & DDRM &\; 29.043  & 0.806  &\; 26.037  & 0.661  &\; 24.341  & 0.551  \\ 
		~ & DDS2M &\; \textbf{36.548}  & \textbf{0.959}  &\; \textbf{32.925}  & \textbf{0.911}  &\; \textbf{30.863}  & \textbf{0.867} \\ \hline
		

	\end{tabular}
	\label{tab:denoising}
\end{table}

\begin{table}[!htbp]
	\scriptsize
	\setlength{\tabcolsep}{3.7pt}
	\renewcommand\arraystretch{1.3}
	\caption{The average quantitative results for noisy HSI super-resolution on CAVE dataset. The \textbf{best} and \underline{second-best} values are highlighted.}
	\centering
	\begin{tabular}{lccccccc}
		\Xhline{1.0pt}
		\multicolumn{2}{c}{Scale} & \multicolumn{2}{c}{$\times$2} &  \multicolumn{2}{c}{$\times$4} & \multicolumn{2}{c}{$\times$8} \\ \hline
		Method & Trained & PSNR & SSIM & PSNR & SSIM & PSNR & SSIM \\ \hline
		SFCSR(0) & \ding{52} &\; 16.615  & 0.198  &\; 16.829  & 0.155  &\; 17.070  & 0.148  \\ 
		SFCSR(0.03) & \ding{52} &\; 24.193  & 0.508  &\; 23.859  & 0.508  &\; 22.277  & 0.472  \\
		SFCSR(0.05) & \ding{52} &\; 27.620  & 0.688  &\; 25.142  & 0.625  &\; 22.953  & 0.587  \\
		SFCSR(0.1) & \ding{52}  &\; 30.350  & 0.856  &\; 26.302  & \underline{0.744}  &\; 23.342  & 0.609  \\
		SFCSR(0.1)* & \ding{52} &\; 28.015  & 0.821  &\; 24.153  & 0.661  &\; 22.011  & 0.569  \\  \hline
		RFSR(0) & \ding{52} &\; 18.570  & 0.252  &\; 18.412  & 0.206  &\; 18.045  & 0.181  \\
		RFSR(0.03) & \ding{52} &\; 26.904  & 0.660  &\; 24.639  & 0.619  &\; 23.023  & 0.576  \\ 
		RFSR(0.05) & \ding{52} &\; 29.591  & 0.814  &\; 26.187  & 0.732  &\; 23.248  & 0.602  \\ 
		RFSR(0.1) & \ding{52} &\; \underline{30.570}  & \textbf{0.868}  &\; 26.479  & \textbf{0.748}  &\; \underline{23.386}  & 0.614  \\ 
		RFSR(0.1)* & \ding{52} &\; 27.994  & 0.804  &\; 24.082  & 0.658  &\; 21.033  & 0.541  \\  \hline
		SSPSR(0) & \ding{52} &\; 18.916  & 0.261  &\; 19.465  & 0.223  &\; 18.636  & 0.204  \\ 
		SSPSR(0.03) & \ding{52} &\; 28.371  & 0.729  &\; 25.351  & 0.654  &\; 22.899  & 0.573  \\ 
		SSPSR(0.05) & \ding{52} &\; 29.799  & 0.830  &\; 26.101  & 0.715  &\; 23.195  & 0.623  \\ 
		SSPSR(0.1) & \ding{52} &\; 30.294  & \textbf{0.868}  &\; \underline{26.824}  & \textbf{0.748}  &\; 23.338  & \textbf{0.627}  \\ 
		SSPSR(0.1)* & \ding{52} &\; 27.636  & 0.828  &\; 24.748  & 0.705  &\; 21.465  & 0.585  \\  \hline
		Bicubic & \ding{56} &\; 21.554  & 0.245  &\; 20.893  & 0.228  &\;  20.021  & 0.238 \\
		LRTV & \ding{56} &\; 20.867  & 0.321  &\; 19.690  & 0.291  &\; 18.490  & 0.280  \\ 
		DIP2D & \ding{56} &\; 28.344  & 0.745  &\; 25.238  & 0.602  &\; 22.613  & 0.482  \\ 
		DIP3D & \ding{56} &\; 27.458  & 0.756  &\; 24.776  & 0.635  &\; 21.935  & 0.506  \\ 
		DDRM & \ding{56} &\; 27.330  & 0.741  &\; 23.244  & 0.552  &\; 18.883  & 0.418  \\ 
		DDS2M & \ding{56} &\; \textbf{30.997}  & \underline{0.859}  &\; \textbf{26.835}  & \textbf{0.748}  &\; \textbf{23.621}  & \underline{0.626} \\ \Xhline{1.0pt}
	\end{tabular}
	\label{tab:superresolution}
\end{table}
\vspace{-0.2cm}
The quantitative results of noisy HSI completion, HSI denoising, and noisy HSI super-resolution are reported in Tables \ref{tab:completion}, \ref{tab:denoising}, and \ref{tab:superresolution}. We can observe that the proposed \texttt{DDS2M} outperforms existing model-based, unsupervised deep learning-based, and diffusion-based methods in all three tasks, while yielding competitive results with respect to the state-of-the-art supervised deep learning-based methods. Specifically, as compared with the diffusion-based method \texttt{DDRM}, our method offers average PSNR improvement of 5.878 dB, 5.909 dB, and 3.998 dB in completion, denoising, and super-resolution, respectively. This observation validates that \texttt{DDS2M} can more flexibly adapt to diverse HSIs in real scenarios. Additionally, in HSI super-resolution experiments, the supervised methods (i.e., \texttt{SFCSR}, \texttt{RFSR}, and \texttt{SSPSR}) all perform best when trained with CAVE dataset with 0.1 Gaussian noise among the five different training settings, and their performance degrades significantly when trained with other noise levels or datasets. It is worth noting that, our \texttt{DDS2M} achieves comparable performance with the best version of these supervised methods, and outperforms the models trained with other settings. This demonstrates the superiority of our \texttt{DDS2M} against these supervised methods.

\begin{figure*}[!htbp]
	\scriptsize
	\setlength{\tabcolsep}{1pt}
	\centering
	\begin{tabular}{cccccccccc}
		Observed & DIP2D & DIP3D & FTNN & FCTN & S2NTNN & HLRTF  & DDRM & DDS2M & Original\\
		
		\includegraphics[width=0.095\textwidth]{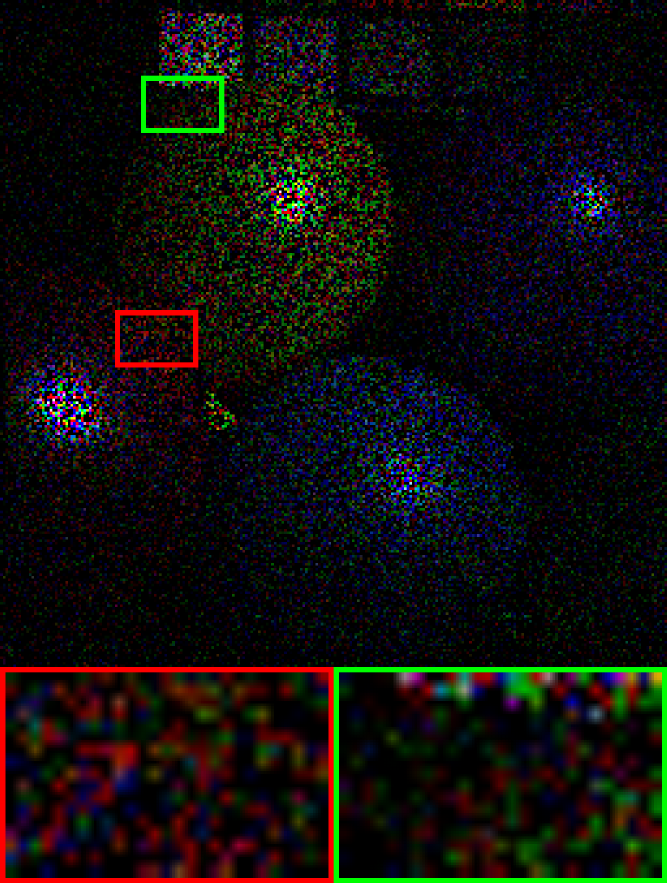}&
		\includegraphics[width=0.095\textwidth]{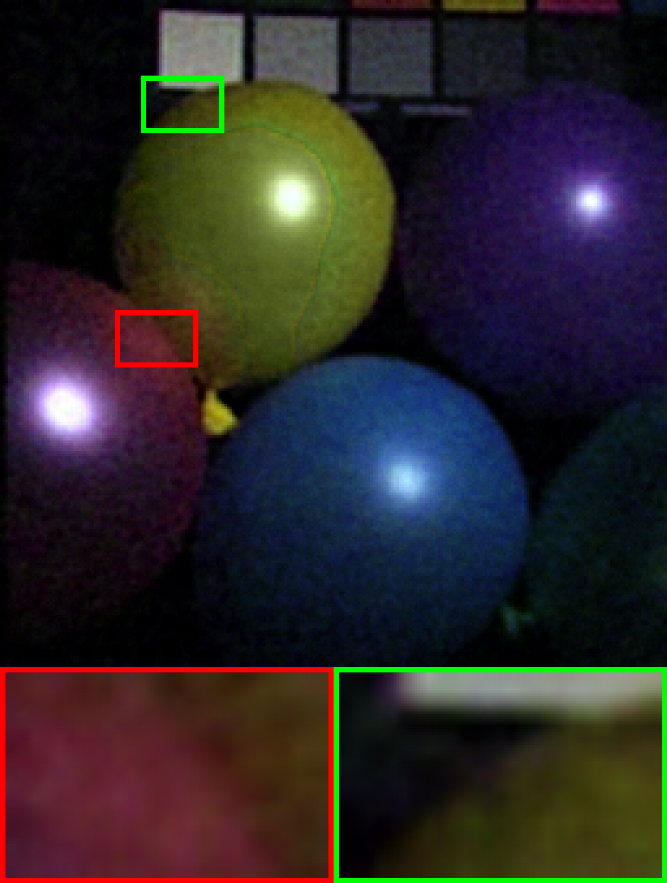}&
		\includegraphics[width=0.095\textwidth]{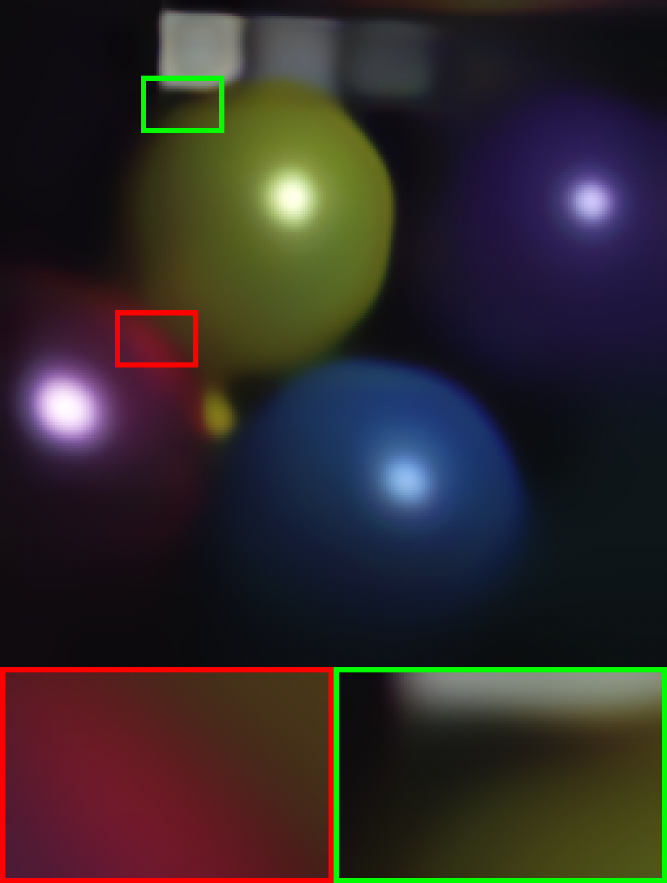}&
		\includegraphics[width=0.095\textwidth]{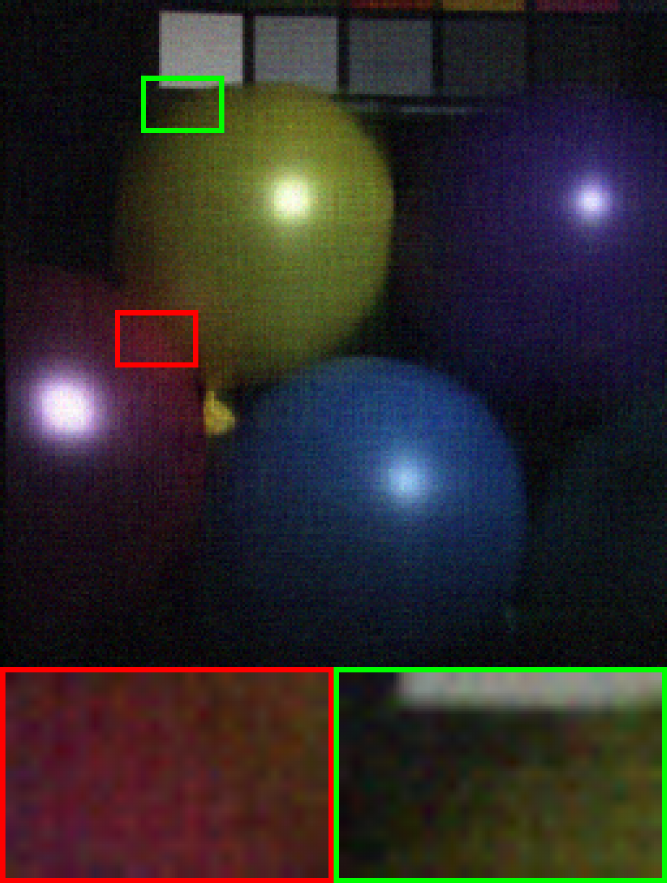}&
		\includegraphics[width=0.095\textwidth]{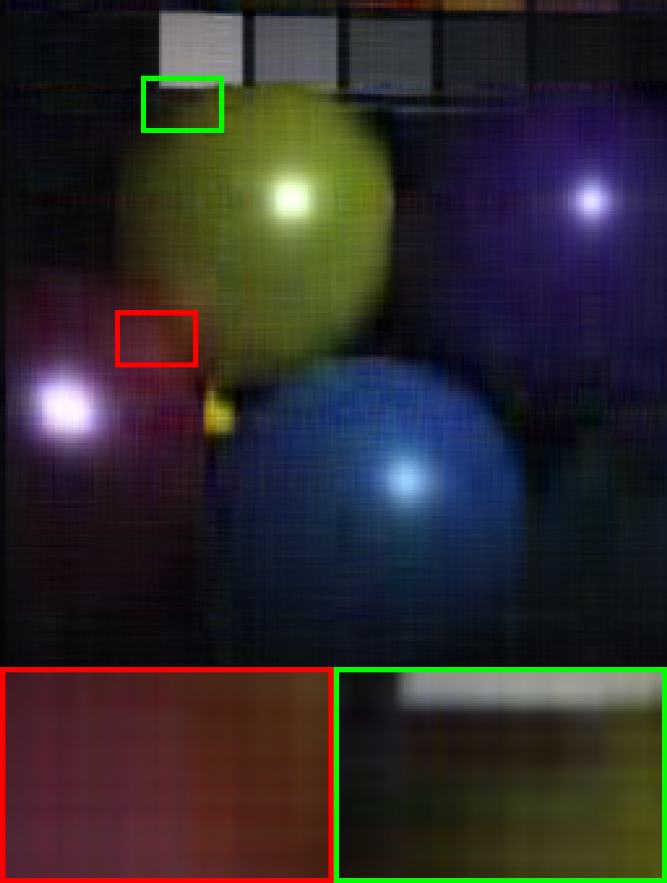}&
		\includegraphics[width=0.095\textwidth]{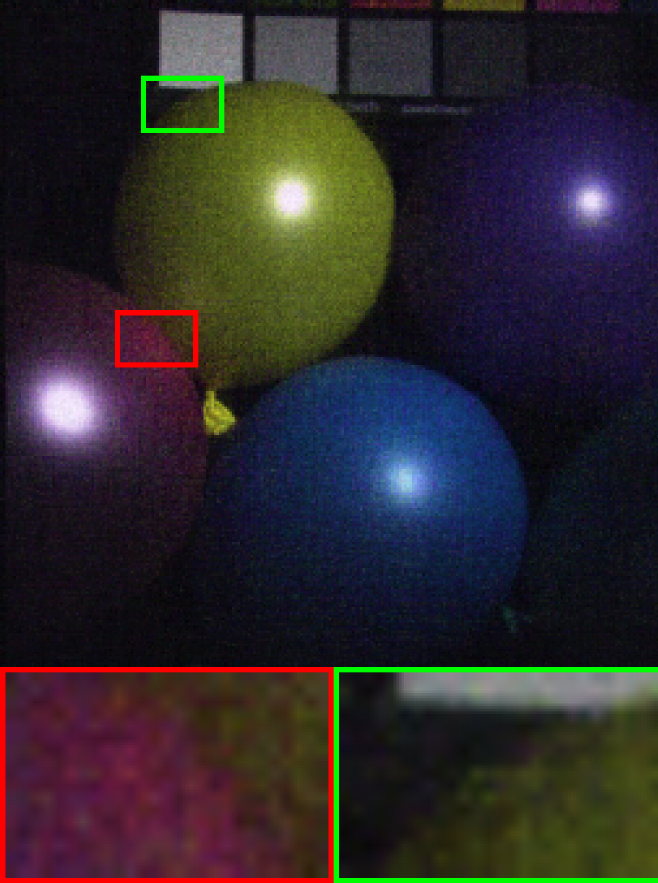}&
		\includegraphics[width=0.095\textwidth]{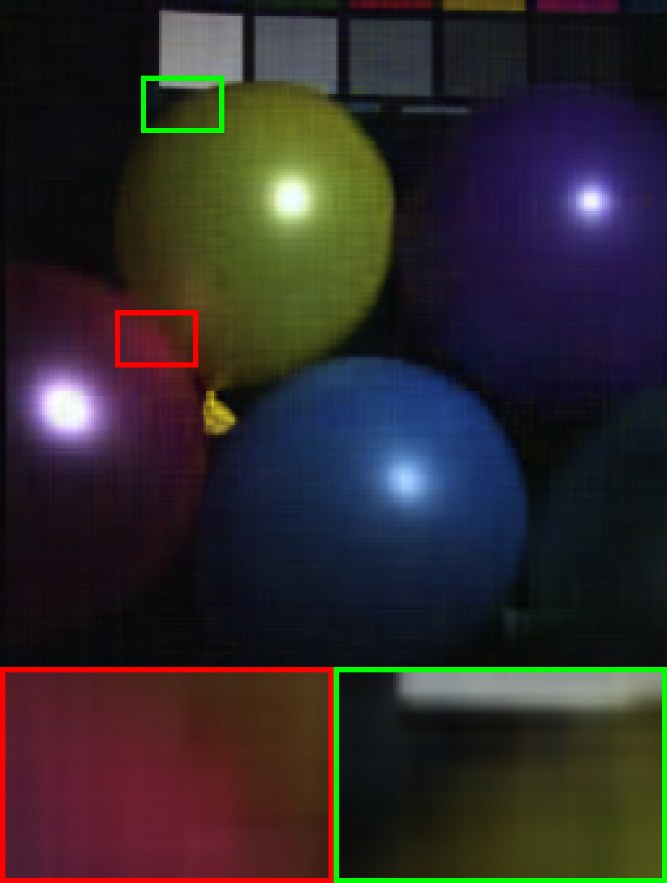}&
		\includegraphics[width=0.095\textwidth]{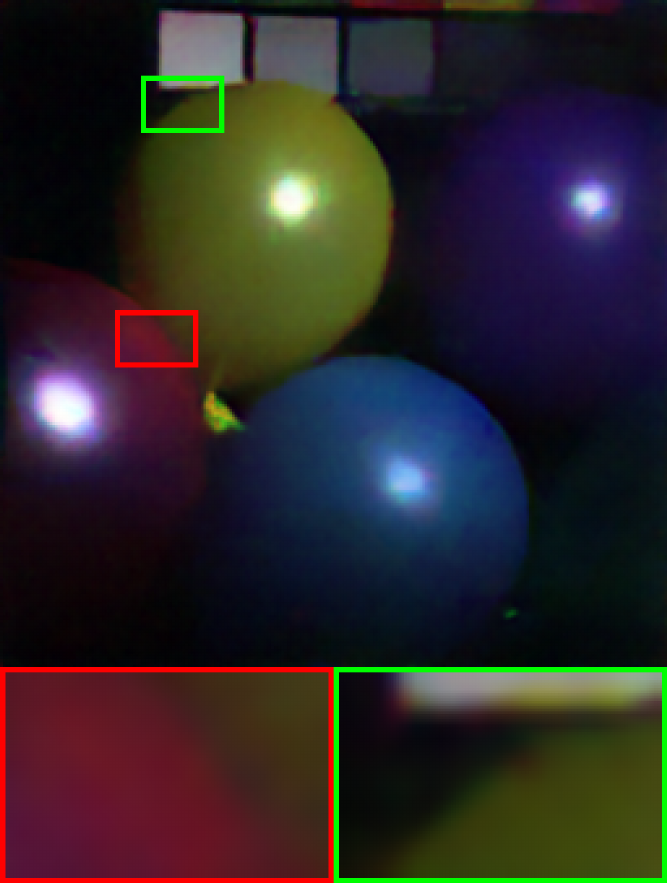}&
		\includegraphics[width=0.095\textwidth]{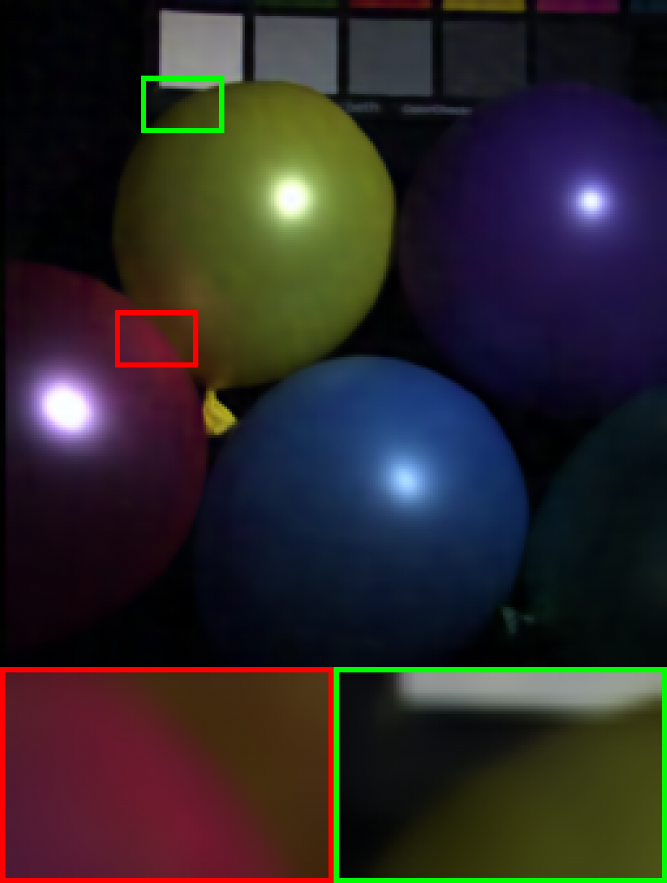}&
		\includegraphics[width=0.095\textwidth]{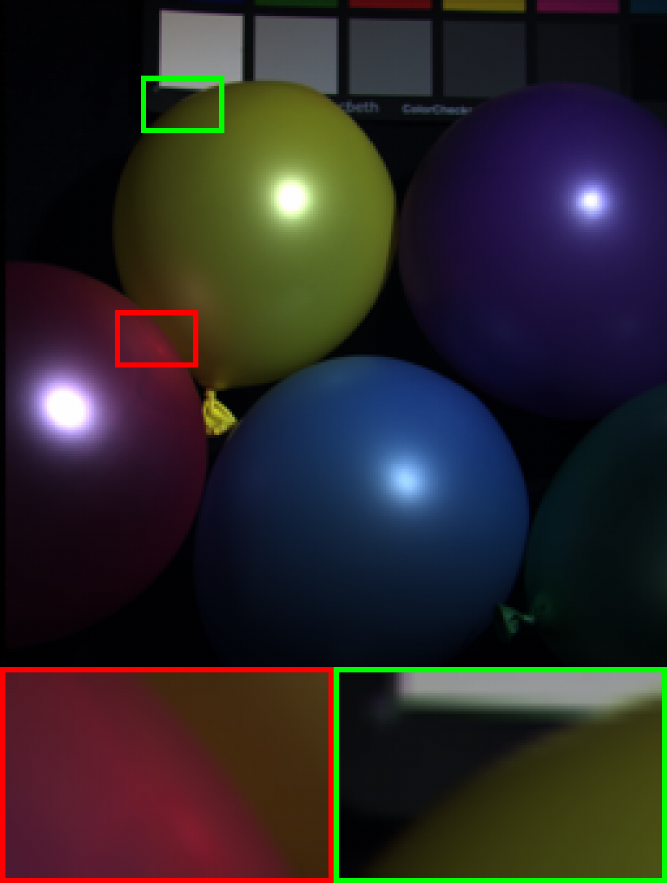}\\
		PSNR 14.582 & PSNR 33.604 & PSNR 28.707 & PSNR 27.877 & PSNR 30.393 & PSNR 28.829 & PSNR 32.778 & PSNR 32.285 & PSNR 37.886 & PSNR Inf \\
		
		
	\end{tabular}
	\caption{The results of noisy HSI completion by different methods on HSI \textit{Balloons} (sampling rate=0.3, $\sigma$=0.1).}
	\label{fig:completion}
\end{figure*}

\begin{figure*}[!htbp]
	\scriptsize
	\setlength{\tabcolsep}{1pt}
	\centering
	\begin{tabular}{cccccccccc}
		Observed & NGMeet & DIP2D & DIP3D & LRTFL0 & E3DTV & HLRTF  & DDRM & DDS2M & Original\\
		
		\includegraphics[width=0.095\textwidth]{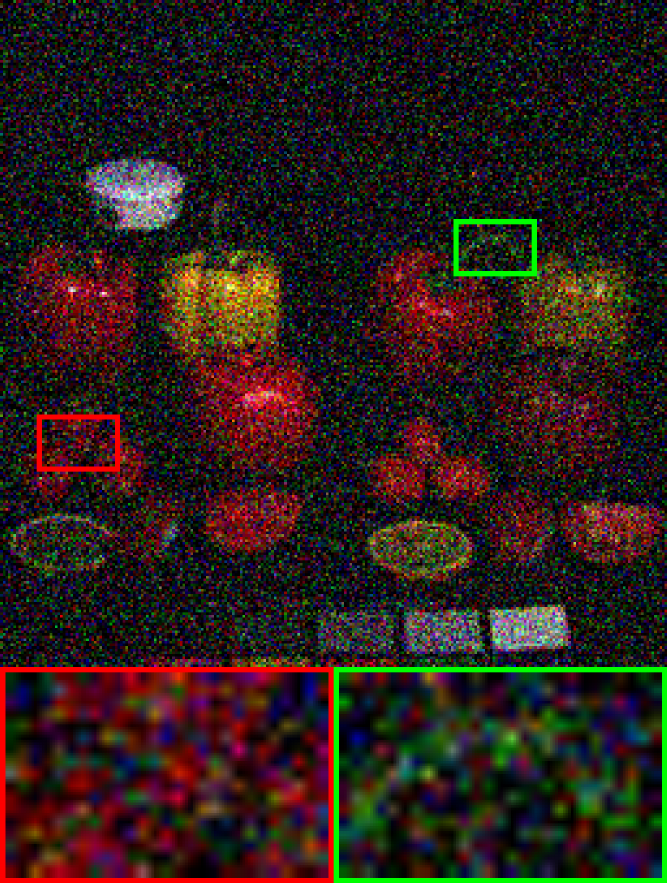}&
		\includegraphics[width=0.095\textwidth]{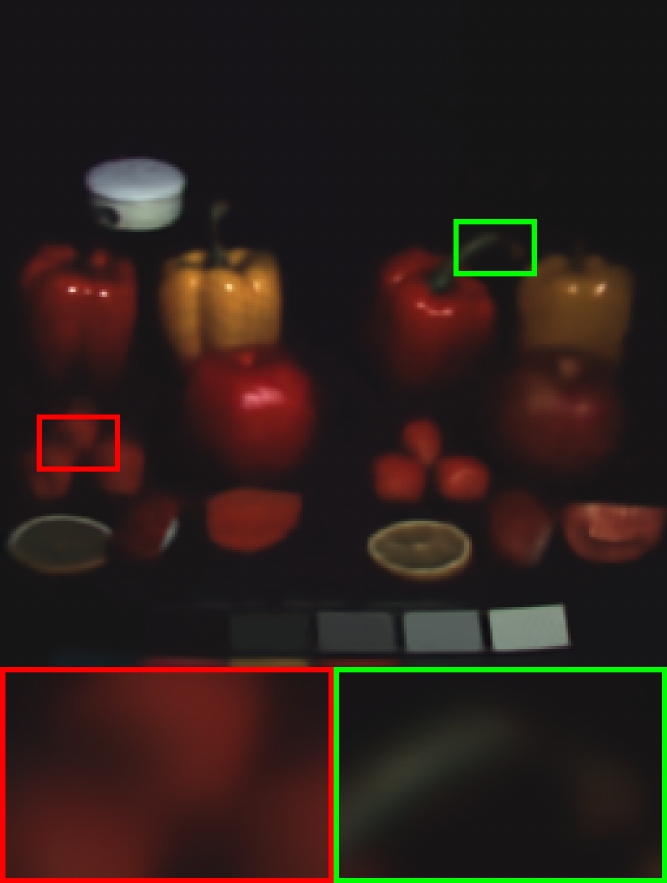}&
		\includegraphics[width=0.095\textwidth]{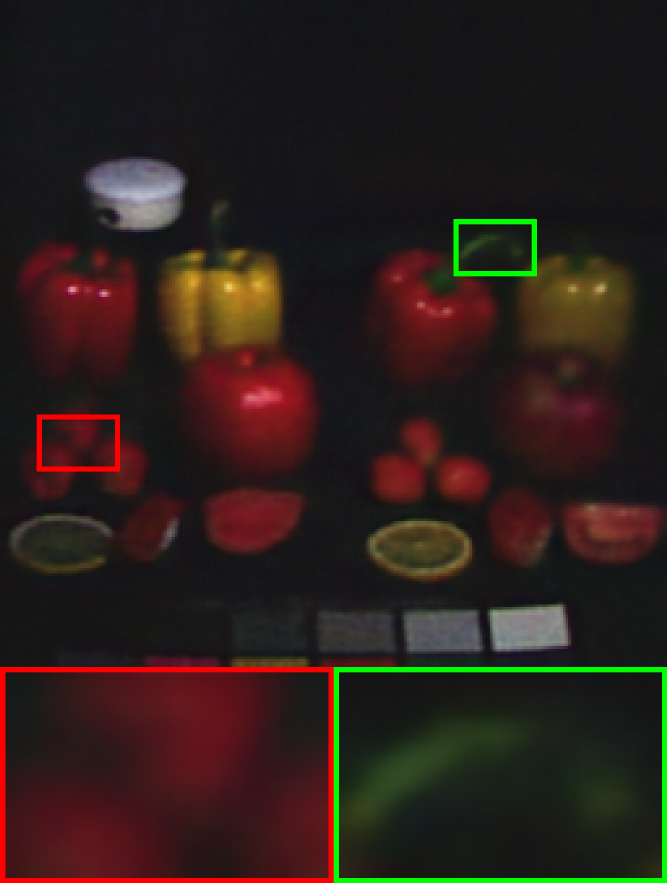}&
		\includegraphics[width=0.095\textwidth]{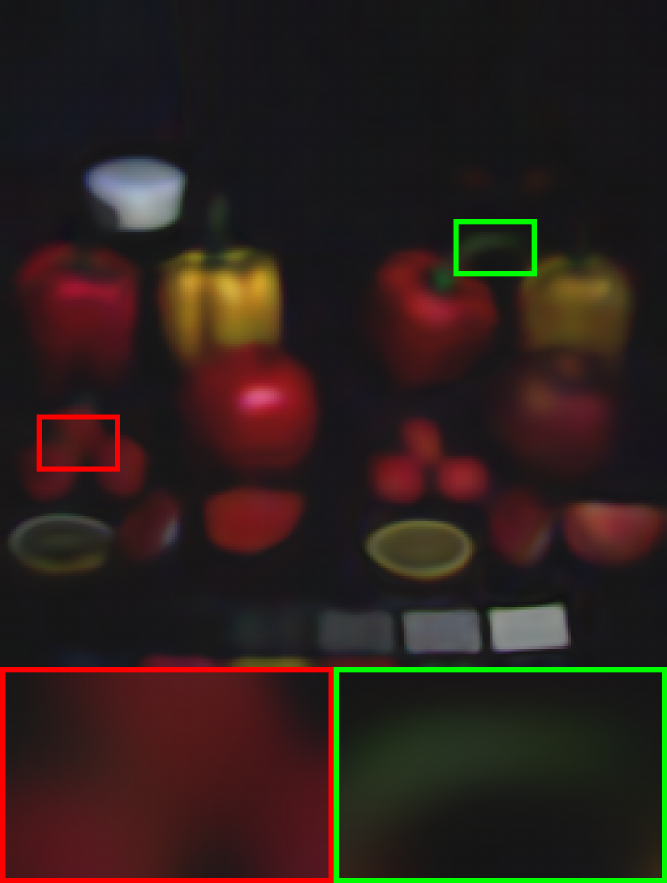}&
		\includegraphics[width=0.095\textwidth]{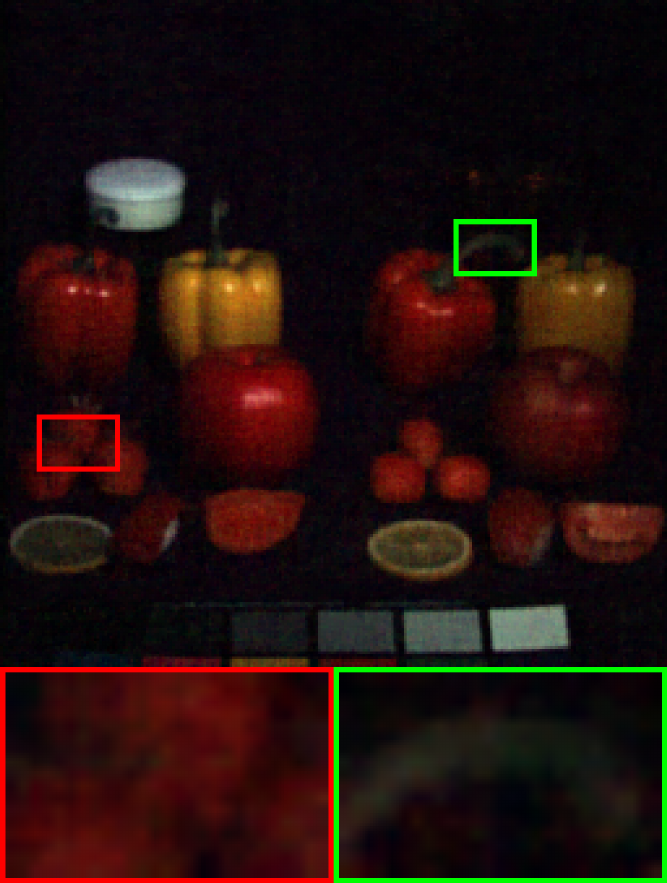}&
		\includegraphics[width=0.095\textwidth]{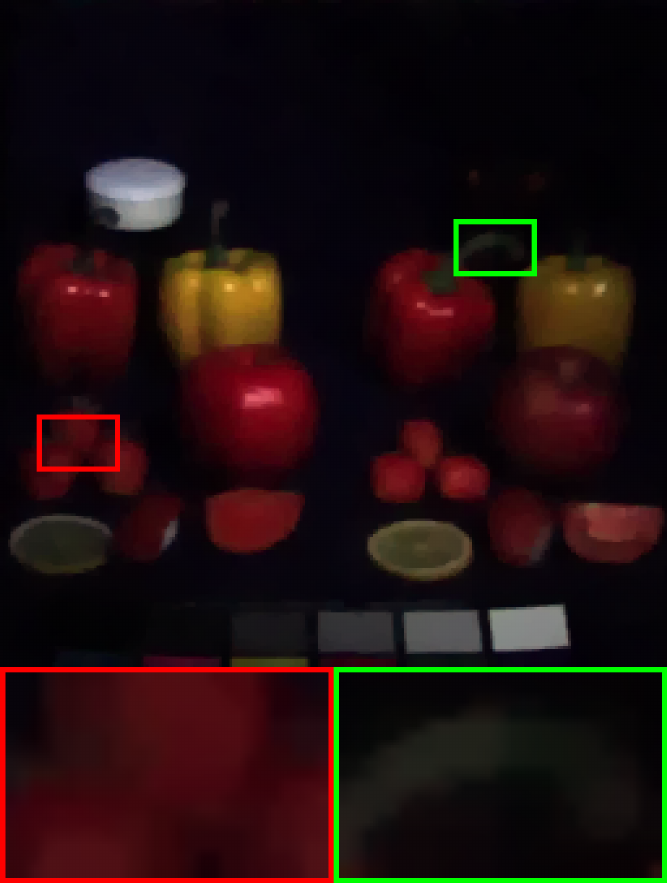}&
		\includegraphics[width=0.095\textwidth]{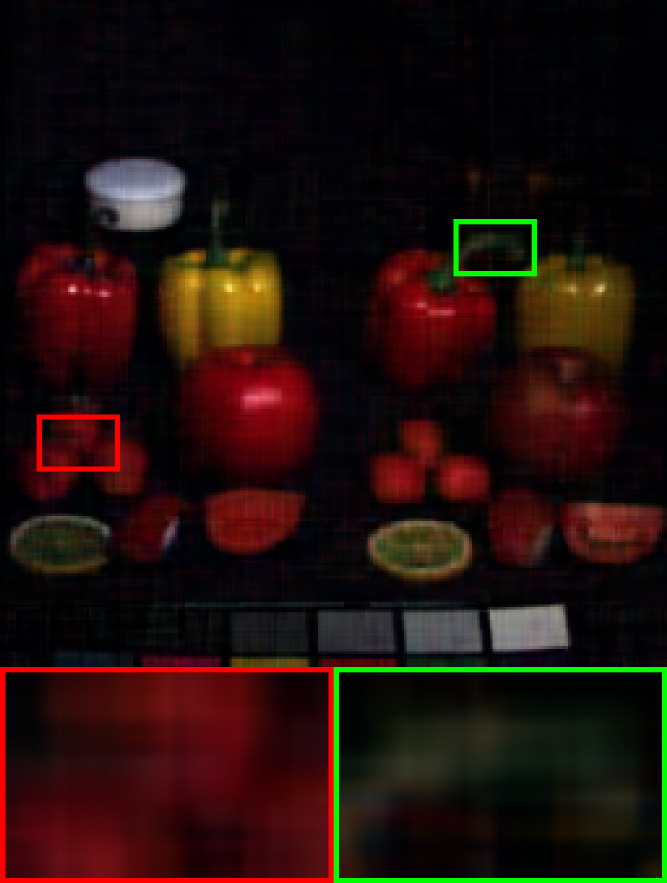}&
		\includegraphics[width=0.095\textwidth]{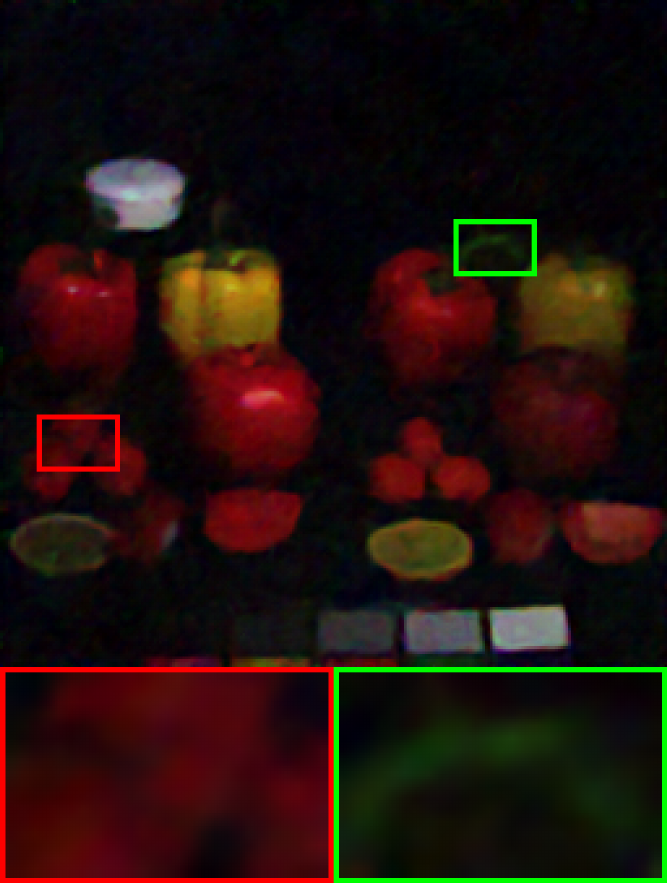}&
		\includegraphics[width=0.095\textwidth]{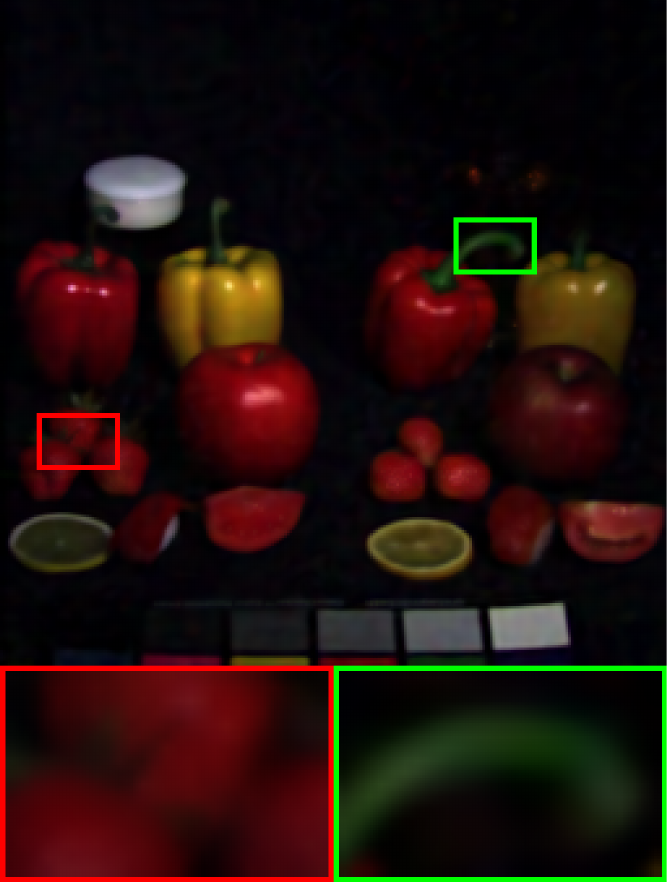}&
		\includegraphics[width=0.095\textwidth]{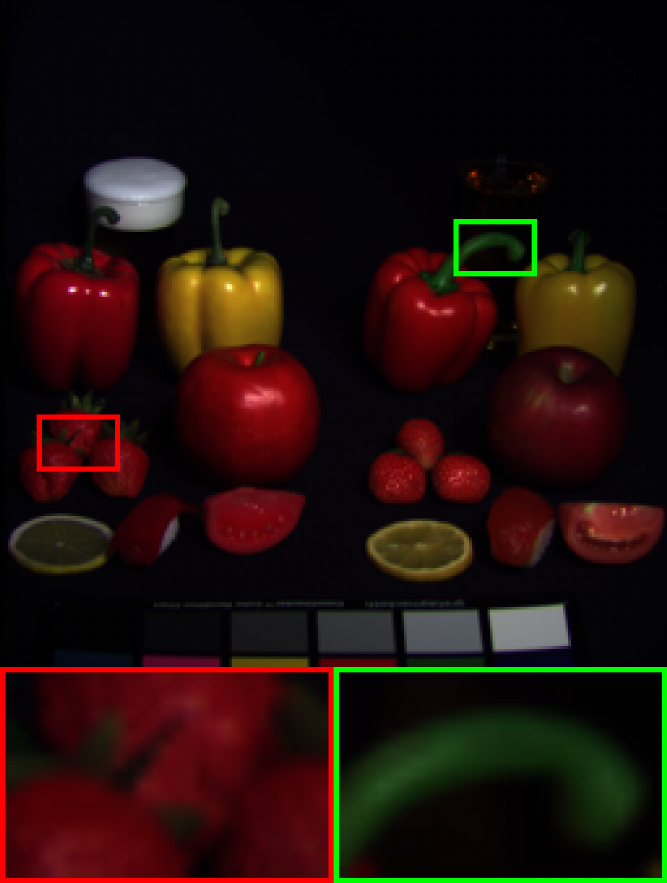}\\
		PSNR 16.423 & PSNR 23.626 & PSNR 23.855 & PSNR 23.087 & PSNR 30.383 & PSNR 32.585 & PSNR 33.658 & PSNR 30.609 & PSNR 34.721 & PSNR Inf \\
		
		
	\end{tabular}
	\caption{The results of HSI denoising by different methods on HSI \textit{Fruits} ($\sigma$=0.2).}
	\label{fig:denoising}
\end{figure*}

\begin{figure*}[!htbp]
	\scriptsize
	\setlength{\tabcolsep}{1pt}
	\centering
	\begin{tabular}{cccccccccc}
		Bicubic & LRTV & DIP2D & DIP3D & SSPSR & SFCSR & RFSR & DDRM & DDS2M & Original\\
		
		\includegraphics[width=0.095\textwidth]{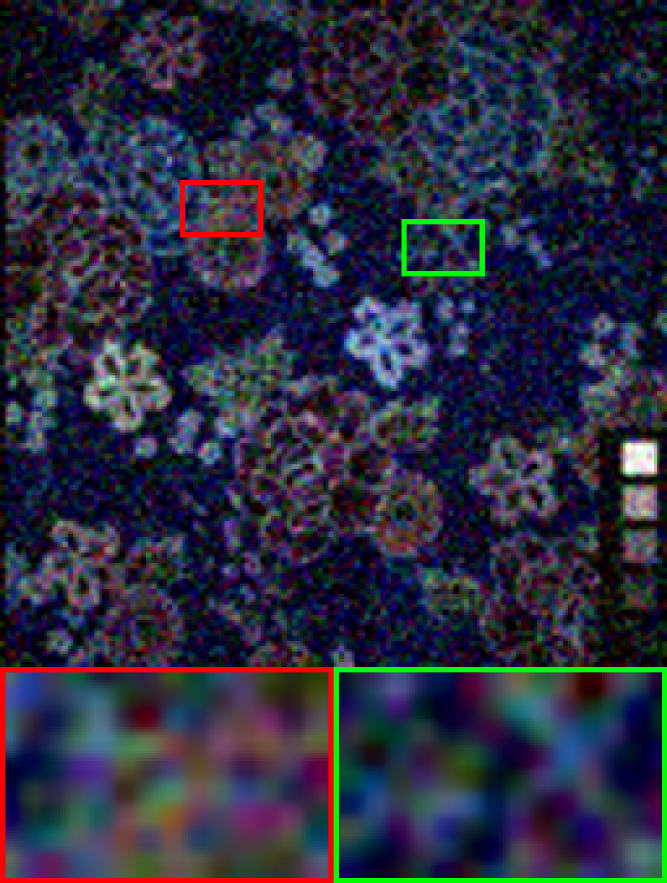}&
		\includegraphics[width=0.095\textwidth]{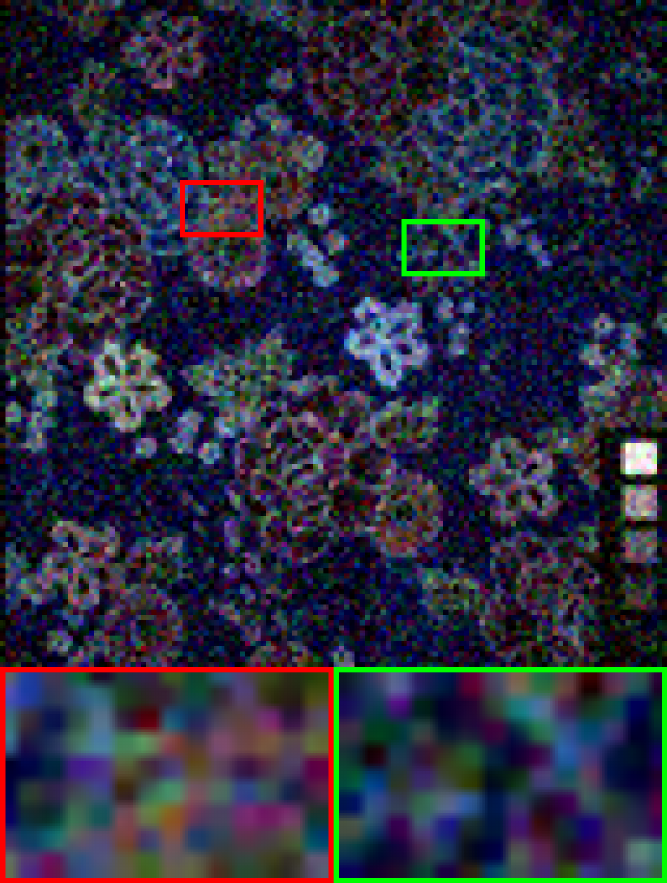}&
		\includegraphics[width=0.095\textwidth]{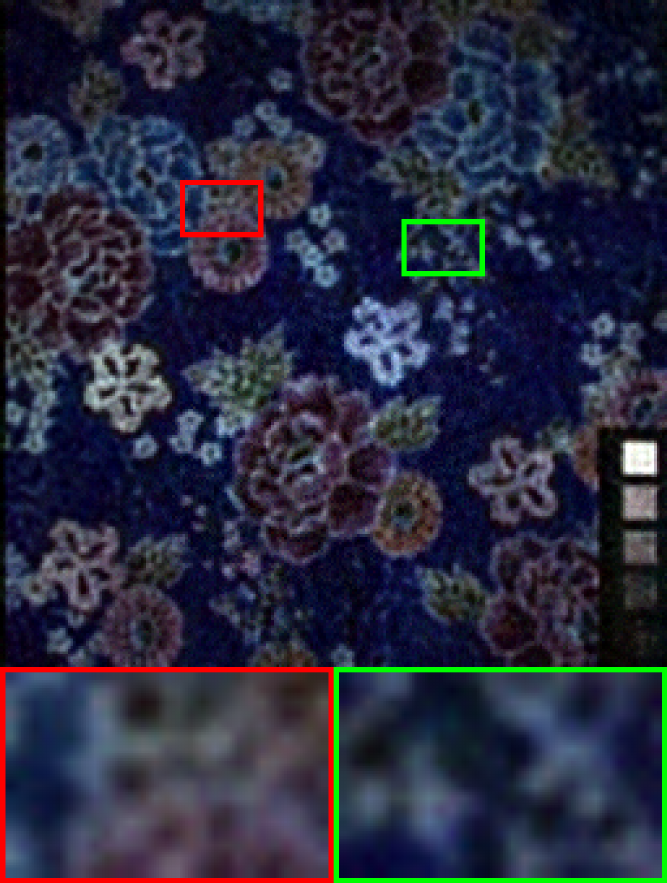}&
		\includegraphics[width=0.095\textwidth]{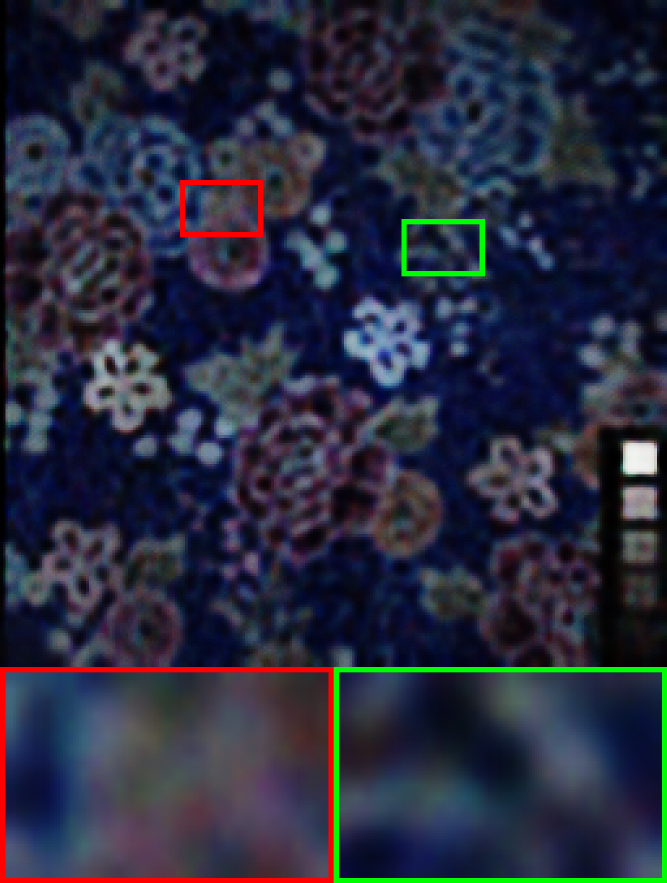}&
		\includegraphics[width=0.095\textwidth]{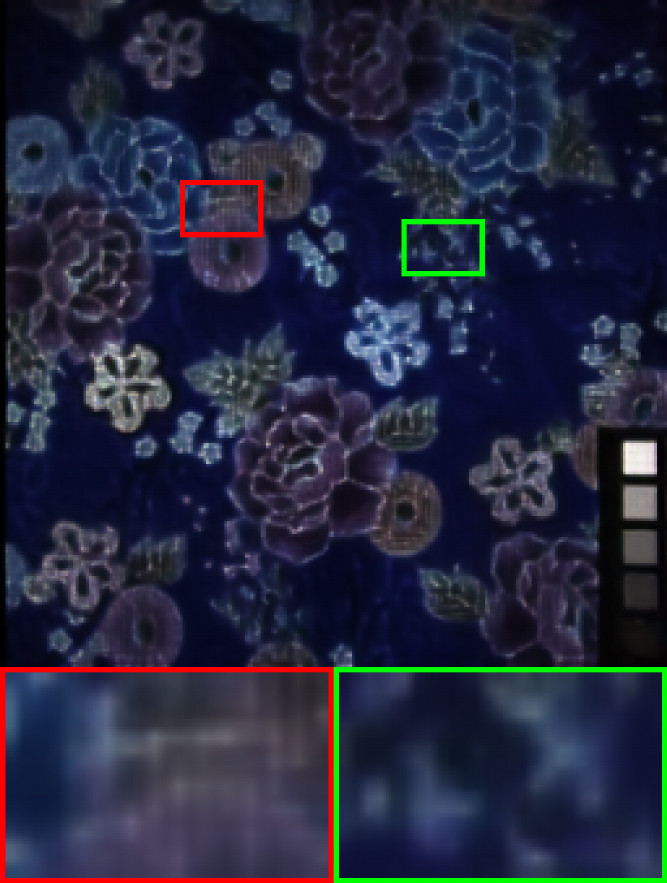}&
		\includegraphics[width=0.095\textwidth]{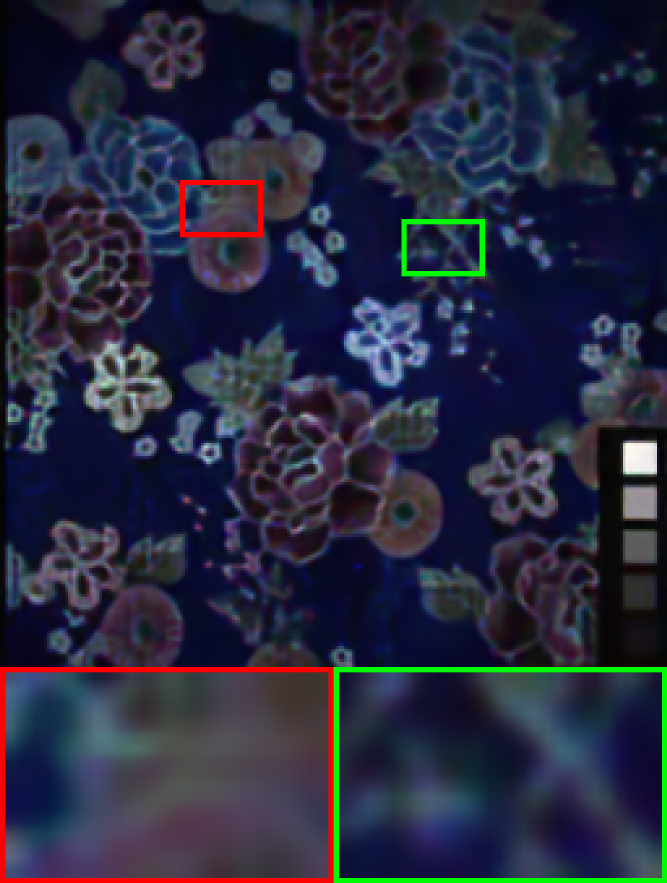}&
		\includegraphics[width=0.095\textwidth]{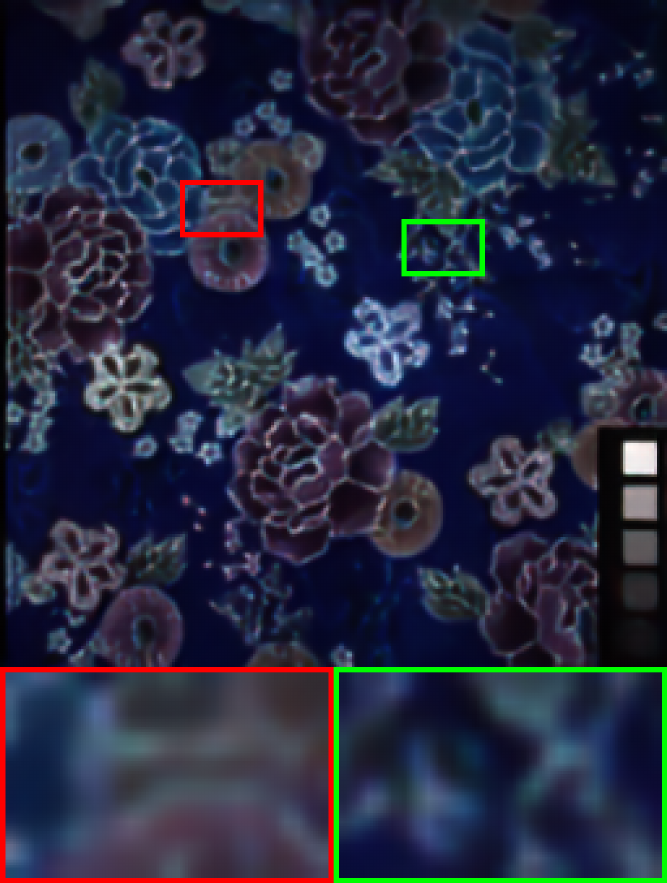}&
		\includegraphics[width=0.095\textwidth]{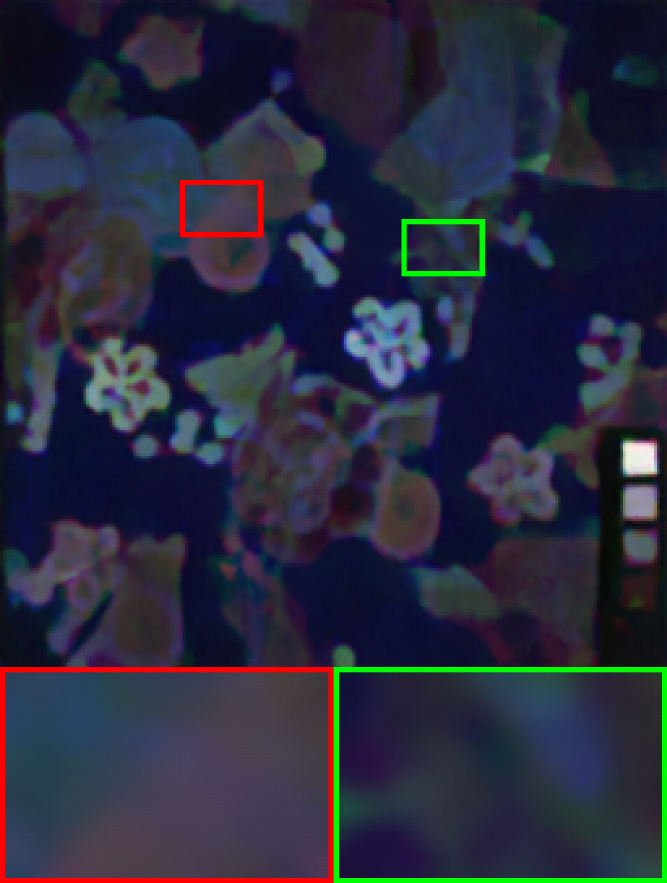}&
		\includegraphics[width=0.095\textwidth]{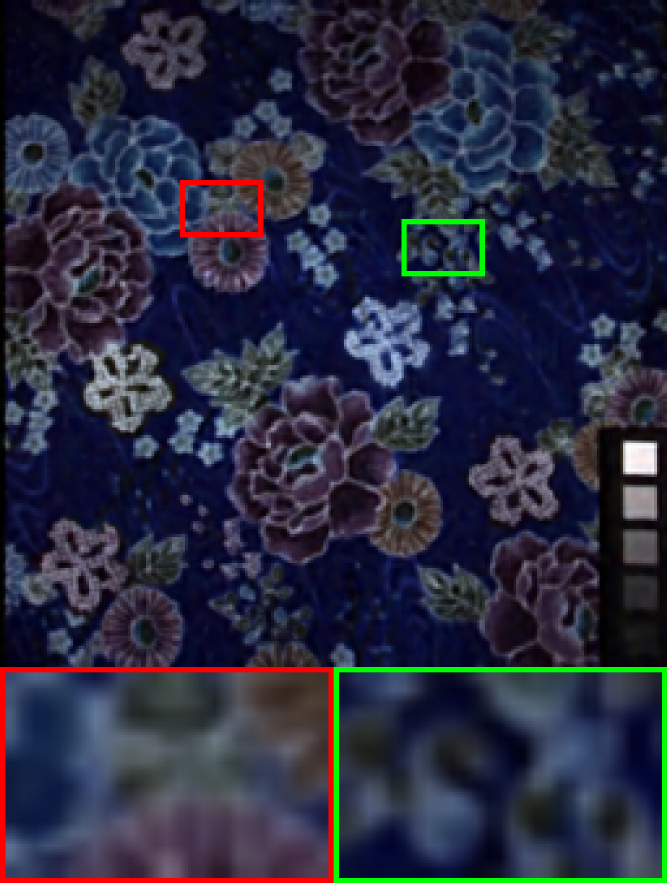}&
		\includegraphics[width=0.095\textwidth]{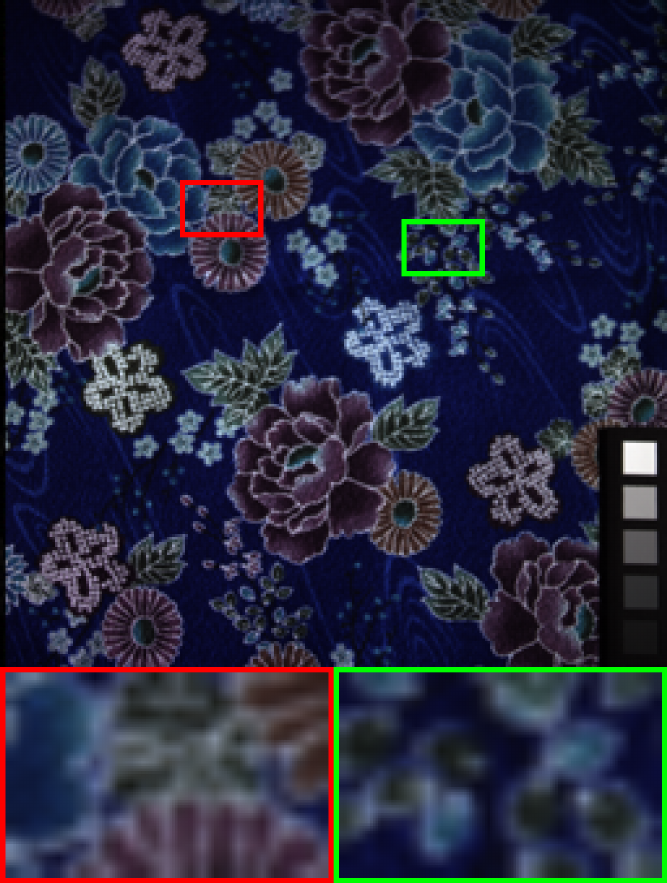}\\
		PSNR 17.875 & PSNR 21.446 & PSNR 24.175 & PSNR 22.991 & PSNR 24.355 & PSNR 24.328 & PSNR 24.575 & PSNR 22.461 & PSNR 25.485 & PSNR Inf \\
		
		
	\end{tabular}
	\caption{The results of noisy HSI super-resolution by different methods on HSI \textit{Cloth} (scale factor=$\times$2, $\sigma$=0.1).}
	\label{fig:super}
\end{figure*}

Some visual results for different tasks are shown in Figures \ref{fig:completion}, \ref{fig:denoising}, and \ref{fig:super}\footnote{In Figure \ref{fig:super} (i.e., super-resolution), the best results of the supervised methods are shown.}.  As observed, the proposed \texttt{DDS2M} is capable of preserving the most detailed information and demonstrating the best visual performance among the compared methods, which is consistent with its satisfactory performance in PSNR and SSIM. In addition, there is the least residual noise remaining in the results produced by \texttt{DDS2M}, which demonstrates the superiority of \texttt{DDS2M} in terms of noise robustness. 

We conjecture that such promising results can be attributed to the organic cooperation of untrained spatial and spectral networks and diffusion model, which is beneficial to the generalization ability to various HSIs and the robustness to noise.

\subsection{Ablation Study}
We test the impact of untrained spatial and spectral networks, and diffusion process in \texttt{DDS2M}. The compared methods are listed as follows:\\
$\bullet$ \textbf{\texttt{DDS2M} without untrained spatial and spectral networks} (dubbed DDS2M w/o untra.): To evaluate the impact of the untrained spatial and spectral networks,  we remove the untrained spatial and spectral networks, and use an untrained U-Net to directly generate the whole HSI.\\
$\bullet$ \textbf{\texttt{DDS2M} without diffusion process} (dubbed DDS2M w/o diffu.):  To clarify the influence of the diffusion process, we remove the diffusion process and make the untrained spatial and spectral networks directly fit the degraded HSI in an iterative scheme.

We consider HSI denoising ($\sigma$=0.3), noisy HSI completion (sampling rate=0.1, $\sigma$=0.1), and noisy HSI super-resolutin (scale factor=2, $\sigma$=0.1). HSI \textit{Fruits} from the CAVE dataset is selected as an example. The results are shown in Table \ref{tab:ablation}. We can observe that the untrained spatial and spectral networks, and the diffusion process could indeed significantly boost the restoration performance. More implementation details can be found in the supplementary materials.


\begin{table}[!ht]
	\scriptsize
	\renewcommand\arraystretch{1.3}
	\setlength{\tabcolsep}{4.5pt}
	\caption{The quantitative ablation results on HSI completion, denoising, and super-resolution. The \textbf{best} values are highlighted.}
	\centering
	\begin{tabular}{ccccccc}
		\Xhline{1.0pt}
		Task & \multicolumn{2}{c}{Denoising} & \multicolumn{2}{c}{Completion} & \multicolumn{2}{c}{Super-Resolution}  \\ \hline
		Methods & PSNR & SSIM & PSNR & SSIM & PSNR & SSIM \\ \hline
		DDS2M w/o diffu. &\; 31.983 & 0.738 &\; 31.562 & 0.739 &\; 31.172 & 0.730 \\ 
		DDS2M w/o untra. &\; 29.682 & 0.643 &\; 28.565 & 0.593 &\; 32.588 & 0.808 \\ 
		DDS2M &\; \textbf{33.045} & \textbf{0.841} &\; \textbf{33.217} & \textbf{0.845} &\; \textbf{34.066} & \textbf{0.876}\\ 
		\Xhline{1.0pt}
	\end{tabular}
	\label{tab:ablation}
\end{table}
\vspace{-0.4cm}
\section{Conclusion}
This work reveals a new insight on how to synergistically integrate existing diffusion models with untrained neural networks, and puts forth a self-supervised diffusion model for HSI restoration, namely Denoising Diffusion Spatio-Spectral Model (\texttt{DDS2M}). By virtue of our proposed Variational Spatio-Spectral Module (VS2M), the diffusion process can be reversed solely using the degraded HSI without any extra training data. Benefiting from its self-supervised nature and diffusion process, \texttt{DDS2M} admits stronger generalization ability to various HSIs relative to existing diffusion-based methods and superior robustness to noise relative to existing HSI restoration methods.


{\small
	\bibliographystyle{ieee_fullname}
	\bibliography{egbib}
}
\end{document}